\pdfoutput=1

\documentclass[11pt]{article}

\usepackage{acl}

\usepackage{times}
\usepackage{latexsym}

\usepackage{amsmath,amsfonts,bm}

\def\eqref#1{equation~\ref{#1}}

\def\1{\bm{1}}

\DeclareMathAlphabet{\mathsfit}{\encodingdefault}{\sfdefault}{m}{sl}
\SetMathAlphabet{\mathsfit}{bold}{\encodingdefault}{\sfdefault}{bx}{n}

\DeclareMathOperator*{\argmin}{arg\,min}

\newcommand\tf[1]{\textbf{#1}}
\newcommand\ttt[1]{\texttt{#1}}

\newcommand{\sent}{\ttt{<}$S_1$\ttt{>}}
\newcommand{\firstsent}{\ttt{<}$S_1$\ttt{>}}
\newcommand{\secondsent}{\ttt{<}$S_2$\ttt{>}}

\newcommand{\mask}{\texttt{[MASK]}}
\renewcommand{\paragraph}[1]{\vspace{0.2cm}\noindent\textbf{#1}}

\usepackage[T1]{fontenc}

\usepackage[utf8]{inputenc}

\usepackage{microtype}

\usepackage{hyperref}
\usepackage{url}
\usepackage{tabularx}

\usepackage{array,makecell}
\usepackage{amssymb}
\usepackage{wrapfig,lipsum,booktabs}
\usepackage{hyperref}
\usepackage{url}
\usepackage{dsfont}
\usepackage{booktabs} 
\usepackage{caption}
\usepackage{amsmath}
\usepackage{mathtools}
\usepackage{sidecap}
\usepackage{multirow}
\usepackage{epsfig,graphics,float}
\usepackage{float}
\usepackage{todonotes}
\usepackage[ruled,vlined]{algorithm2e}
\usepackage{subfig,balance,lineno}
\usepackage{float}
\usepackage{verbatim}

\usepackage{wrapfig}
\usepackage{amssymb}
\usepackage{arydshln}
\usepackage{tikz}
\usepackage{pgfplots}
\usepackage[normalem]{ulem}
\usepackage{xcolor}
\usepackage{dashbox}%
\usepackage{multirow}
\usepackage{textcomp}
\usepackage{wrapfig}

\DeclarePairedDelimiterX{\infdivx}[2]{(}{)}{%
  #1\;\delimsize\|\;#2%
}
\newcommand{\infdiv}{\mathrm{KL}\infdivx}

\newcommand{\sysname}{{\tt LiST}}

\title{LiST: Lite Prompted Self-training Makes Parameter-efficient\\ Few-shot Learners}

\author{Yaqing Wang$^{\S}$\thanks{~Most of the work was conducted while the first author was interning at Microsoft.}, Subhabrata Mukherjee$^\dagger{}$, Xiaodong Liu$^\dagger{}$, \\\textbf{Jing Gao}$^{\S}$, \textbf{Ahmed Hassan Awadallah}$^\dagger{}$, \textbf{Jianfeng Gao}$^\dagger{}$ \\
  $^{\S}$Purdue University,
  $^\dagger{}$Microsoft Research\\
  \texttt{\{wang5075, jinggao\}@purdue.edu},\\ \texttt{\{submukhe, xiaodl, hassanam, jfgao\}@microsoft.com}}

\begin{document}
\maketitle
\begin{abstract}
We present a new method {\tt LiST}\footnote{{\sysname} is short for \textbf{Li}te Prompted \textbf{S}elf-\textbf{T}raining.} for parameter-efficient fine-tuning of large pre-trained language models (PLMs) for few-shot learning. {\sysname} improves over recent methods that adopt prompt-based fine-tuning (FN) using two key techniques. The first is the use of self-training to leverage large amounts of unlabeled data for prompt-based FN in few-shot settings. We use self-training in conjunction with meta-learning for re-weighting noisy pseudo-prompt labels. Self-training is expensive as it requires updating all the model parameters repetitively. Therefore, we use a second technique for light-weight fine-tuning where 
we introduce a small number of task-specific parameters that are fine-tuned during self-training while keeping the PLM encoder frozen. 
 Our experiments show that {\sysname} can effectively leverage unlabeled data to improve the model performance for few-shot learning. Additionally, the fine-tuning is efficient as it only updates a small percentage of parameters and the overall model footprint is reduced since several tasks can share a common PLM encoder as backbone. A comprehensive study on six NLU tasks demonstrate {\sysname}\url{} to improve by $35\%$ over classic fine-tuning and $6\%$ over prompt-based FN with $96\%$ reduction in number of trainable parameters when fine-tuned with no more than $30$ labeled examples from each task. With only $14M$ tunable parameters, {\sysname} outperforms GPT-3 in-context learning by $33\%$ on few-shot NLU tasks\footnote{Our
code is publicly available at \url{https://github.com/microsoft/LiST}}. 
\end{abstract}
          \section{Introduction}

\begin{figure}[hbt!]

\subfloat[Performance Comparison]{
\begin{minipage}{0.8\linewidth}
\includegraphics[width=1.0\linewidth]{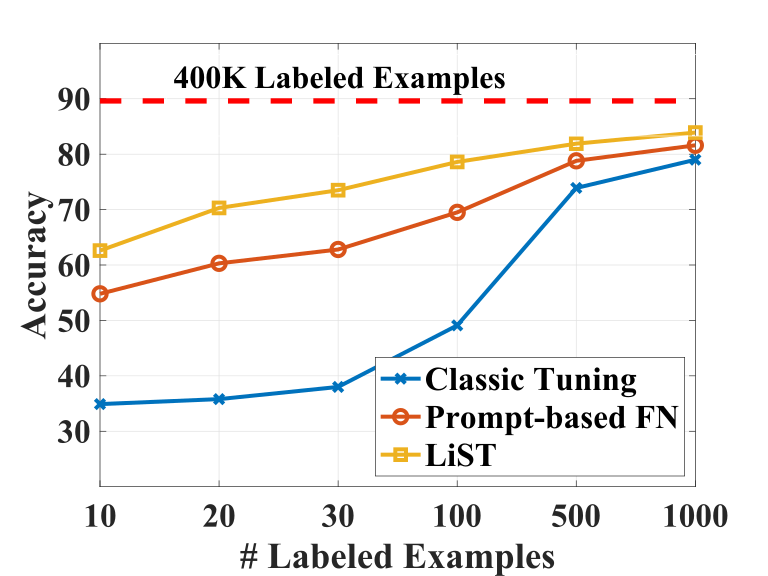}\vspace{-0.15in}
\label{subfig:mnli}
\end{minipage}}

\subfloat[Tunable Parameters]{
\begin{minipage}{0.8\linewidth}
\includegraphics[width=1.0\linewidth]{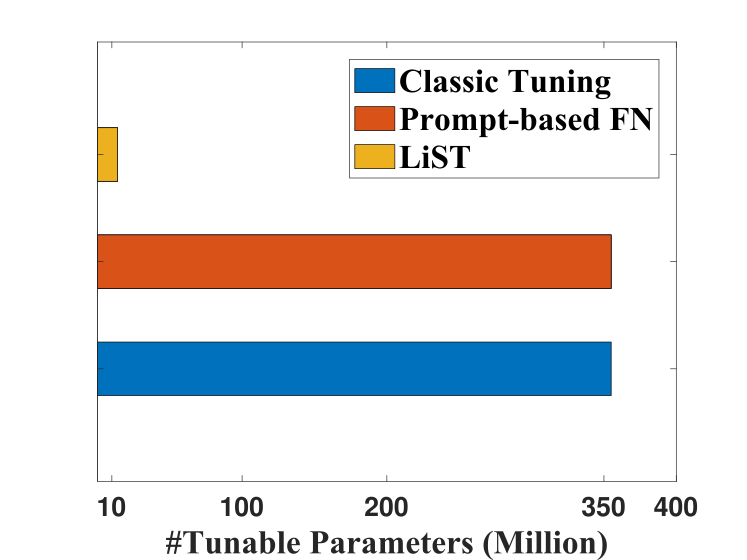}\vspace{-0.15in}
\label{subfig:rte}
\end{minipage}}

 \caption{{\sysname} leverages prompt-based fine-tuning (FN) with unlabeled data for label-efficiency and adapters for reducing tunable parameters. (a) shows classic tuning, prompt-based FN and {\sysname} using RoBERTa-large as backbone on MNLI task for a comparison. The red dash line depicts ceiling performance with full supervision  with RoBERTa-large.  (b) shows the number of tunable parameters for each method. Table~\ref{tab_app:gpt-3} shows a comparison with GPT-3 in-context learning. }\label{fig:key-result}
 \vspace{-0.1in}
\end{figure}

Large pre-trained language models (PLMs) have obtained state-of-the-art performance in several natural language understanding tasks~\citep{DBLP:conf/naacl/DevlinCLT19,DBLP:conf/iclr/ClarkLLM20,DBLP:journals/corr/abs-1907-11692}. Despite their remarkable success in fully supervised settings, their performance is still not satisfactory when fine-tuning with only a handful of labeled examples. While models like GPT-3~\citep{brown2020language} have obtained impressive few-shot performance with in-context task adaptation, they have a significant performance gap relative to fully supervised SoTA models. For instance, few-shot GPT-3 performance is $20$ points worse than fully-tuned DeBERTa~\citep{DBLP:conf/iclr/HeLGC21} on SuperGLUE. This poses significant challenges for many real-world tasks where large labeled data is difficult to obtain.



In this work, we present a new fine-tuning method {\sysname} that aims to improve few-shot learning ability over existing fine-tuning strategies using two techniques as follows.

 The first one is to leverage self-training with large amounts of unlabeled data from the target domain to improve model adaptation in few-shot settings. Prompt-based fine-tuning~\citep{gao2021making} have recently shown significant improvements over classic fine-tuning in the few-shot learning setting. In this paper, we demonstrate that self-training with unlabeled data is able to significantly improve prompt-based fine-tuning~\citep{gao2021making} where we iteratively update a pair of teacher and student models given {\em natural language prompts} and {\em very few labeled examples for the task}. Since the uncertain teacher in few-shot setting produces noisy pseudo-labels, we further use meta-learning to re-weight the pseudo-prompt labels.

 Traditional self-training can be expensive if we have to update all model parameters iteratively. To improve the efficiency of self-training, the second key technique introduces a small number of task-specific adapter parameters in the PLM that are updated with the above technique, while keeping the large PLM encoder fixed. We demonstrate such light-weight tuning with self-training to match the model performance where all parameters are tuned.  
This enables {\em parameter-efficient use of self-training} and {\em reduces the storage cost} of the fine-tuned model since multiple fine-tuned models can now share the same PLM as backbone during inference. Note that the computational cost of inference is out of the scope of this paper and previous work has studied several ways to address it including techniques like model distillation~\citep{hinton2015distilling} and pruning~\citep{lecun1990optimal}.

We perform extensive experiments in six natural language understanding tasks to demonstrate the effectiveness of {\sysname}. We devise a comprehensive evaluation framework considering the variance in few-shot performance of PLMs with different shots, random seeds and splits\footnote{\href{https://github.com/microsoft/List}{LiST} repository contains the dataset partitions for different shots, seeds and splits for every task for reproducibility and benchmarking of efficient few-shot language models.}. Results show that {\sysname} improves over traditional and more recent prompt-based FN  methods by $35\%$ and $6\%$, respectively, with $96\%$ reduction in number of trainable parameters given only $30$ labeled examples for each downstream task. Figure~\ref{fig:key-result} shows the results on MNLI \citep{mnli} as an example. We compare {\sysname} with GPT-3 in-context learning outperforming it by $33\%$ as well as several SoTA few-shot  semi-supervised learning approaches with {\sysname} outperforming the strongest baseline by $6\%$ given only $30$ labeled examples for each task.

\noindent {\bf  Problem statement.} Each downstream task in our framework consists of very few labeled training examples $\mathcal{D}^{Train}_K$ for different shots $K \in \{10, 20, 30\}$ where $|\mathcal{D}^{Train}_K| = K$, unlabeled data $\mathcal{D}^U$ where $\mathcal{D}^{U} \gg \mathcal{D}^{Train}_K$, and a test set $\mathcal{D}^{Test}$.




Given above dataset $\mathcal{D}_K=\mathcal{D}^{Train}_K \cup \mathcal{D}^U$ for a task with shots $K$, a PLM with parameters $\Theta_{\mathrm{PLM}}$ and loss function $\mathcal{L}$, we want to adapt the model for the few-shot learning task by introducing a small number of tunable model parameters $\psi \ll \Theta_{\mathrm{PLM}}$. 

\section{Background on Model Fine-tuning}
\label{subsec:primer}

Given a text sequence ${x}$ or a pair of sequences $\{x_1, x_2\}$ separated by special operators (e.g., {\small [CLS]} and {\small [SEP]}) and a language model encoder $enc (\theta)$ parameterized by $\theta$ -- classic fine-tuning popularized by \citep{BERT} leverages hidden state representation $h_{[CLS]}$ of the sequence(s) obtained from $enc([\text{\small [CLS]}\ x_1\ \text{\small [SEP]}\ x_2\ \text{\small [SEP]}])$ as input to a task-specific head $softmax(W^T \cdot h_{[CLS]})$ for classification, where $W \in \mathbb{R}^{d \times L}$ with $d$ and $L$ representing the hidden state dimension and number of classes, are randomly initialized tunable parameters. In the process it updates both task-specific head $W$ and encoder $\theta$ parameters jointly.



\begin{figure}[hbt!]
\centering
\includegraphics[width=0.5\linewidth]{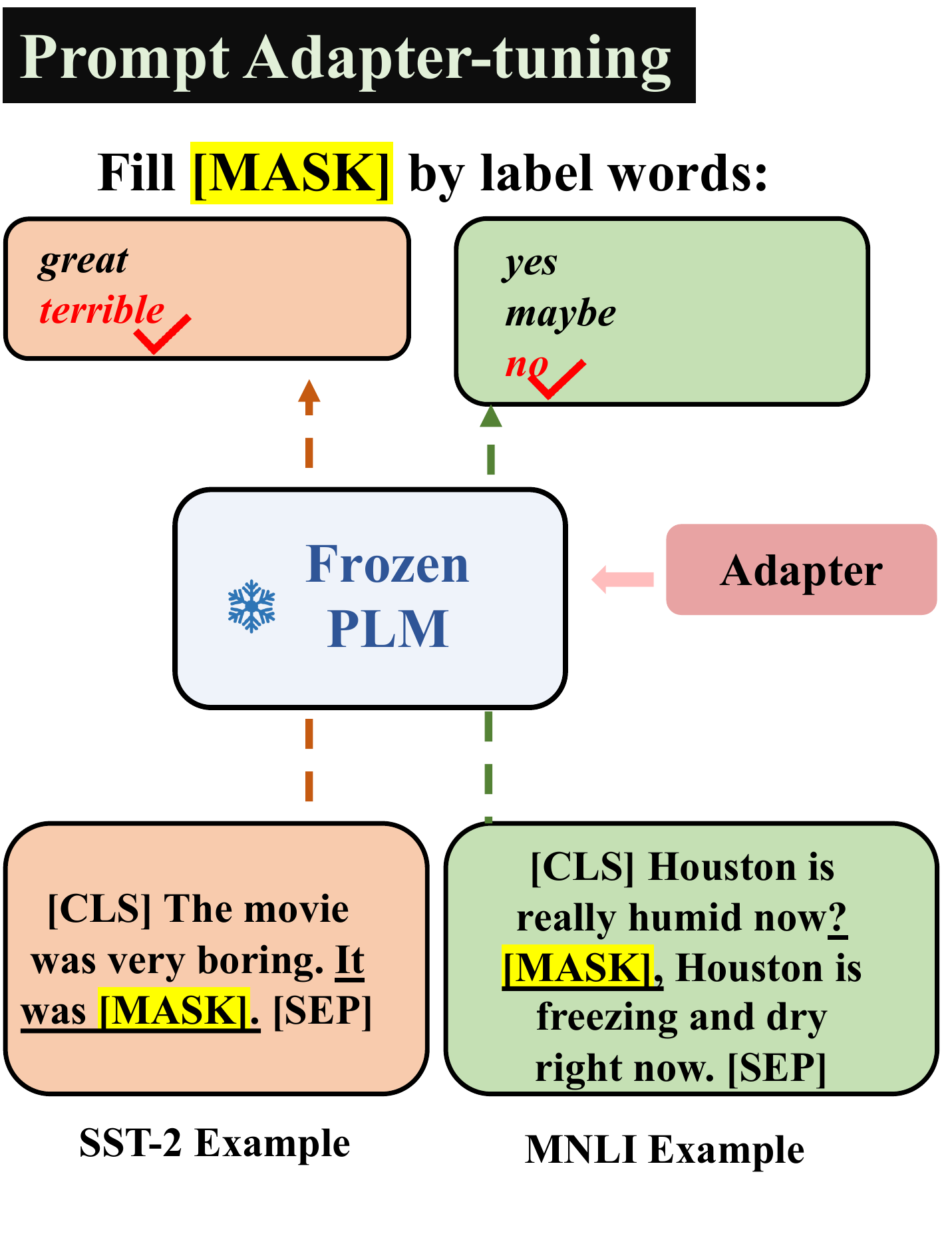}
\vspace{-0.1in}
\caption{The underlined text depicts task prompt to transform classification into Fill-in-MASK task. Label words are  used as proxy for original task labels.}\label{fig:prompt}
\vspace{-0.1in}
\end{figure}

However, this introduces a gap between pre-training and fine-tuning objective with disparate label spaces and additional randomly initiated parameters $W$ introduced for task-specific fine-tuning. This is particularly challenging for few-shot classic fine-tuning, where the limited labeled data is inadequate for adapting the task-specific head and PLM weights effectively. 
Prompt-based FN~\citep{schick-schutze-2021-just, gao2021making} addresses this gap, by re-formulating the objective as a cloze-style auto-complete task. This is done by adding a phrase (also called {\em prompt}) to a sentence like $x_1 =\ $``{\tt \small contains no wit, only labored gags}" in the form of $\Tilde{x} = x_1\ \oplus $ ``{\tt \small It was [MASK]}", where $\oplus$ denotes concatenation of two strings; and output mappings (also called {\em verbalizers}) from vocabulary $\mathcal{V}$ to the label space $\mathcal{Y}$ like ``{\tt \small \{great, terrible\}}" corresponding to positive and negative classes (refer to Figure~\ref{fig:prompt} for an example).
The probability of predicting class $y \in \mathcal{Y}$ is equal to calculating the probability of corresponding label word $v \in \mathcal{V}$:
\begin{equation}
\vspace{-0.1in}
\label{eq:prompt-tune}
\small
p(\mathrm{[MASK]} = v |\Tilde{x}) = \frac{\exp(W^T_v \cdot h_{\mathrm{[MASK]}})}{\sum_{v' \in V} \exp(W^T_{v'} \cdot h_{\mathrm{[MASK]}})}
\end{equation}
\noindent where  $W_v$ indicates the tunable parameters. Since it is identical to masked language modeling (MLM), $W_v$ is initialized by pre-trained weights of PLMs.






{\em In this work, we demonstrate lite self-training with unlabeled data to significantly improve prompt fine-tuning of PLMs in few-shot settings.}




\section{Related Works}
\vspace{-0.06in}
{\noindent\textbf{Few-shot and Semi-supervised Learning.} Recent works have explored semi-supervised methods for few-shot learning with task-specific unlabeled data, including data augmentation~\citep{xie2019unsupervised,   vu2021strata}, self-training~\citep{he2019revisiting, mukherjee2020uncertainty, wang2021meta} and contrastive learning~\citep{gunel2020supervised}. GPT-3~\citep{brown2020language} leverages massive scale with 175 billion parameters to obtain remarkable few-shot performance on several NLU tasks given {\em natural language prompts} and a few {\em demonstrations} for the task. Recent works~\citep{schick2021exploiting, gao2021making} extend this idea of {\em prompting} to language models like BERT~\citep{BERT} and RoBERTa~\citep{RoBERTa}. The most related work to ours is iPET~\citep{schick2021exploiting}, which combines prompt-based FN  with semi-supervised learning. While iPET ensembles multiple fully-tuned models, we develop a lightweight prompted self-training framework to achieve both data and parameter efficiency. 


Few-shot adaptation works~\citep{finn2017model, li2019learning,zhong2021useradapter,wang2021multimodal, wang2021learning, beck2021adapterhub} train models on {\em massive labeled data on source tasks} and develop techniques to adapt them to {\em a target task with few-shot labels}. For instance, \cite{beck2021adapterhub} first trains on miniImagenet with {\em 38,400 labeled examples} (64 classes and 600 samples per class). Similarly, ~\cite{zhong2021useradapter,beck2021adapterhub} first train their models on thousands of labels from the source task to study few-shot target adaptation. In contrast, we focus on \textbf{\em single-task true few-shot learning} with only \textbf{10 to 30} labeled examples available overall and \textbf{no auxiliary supervision labels}. Refer to \cite{perez2021true} for an overview of true few-shot learning for NLU. The objective of few-shot adaptation and true few-shot learning is also quite different. The objective of true few-shot learning is to learn a new task with limited labeled data while the objective of few-shot adaptation is to efficiently transfer to a new task/domain with limited labeled data. Thus, few-shot adaptation leverages multi-task setting with auxiliary labeled data from the source tasks that are not available in our setting. The most relevant works to our setup~\citep{gao2021making,wang2021meta,schick2021exploiting,mukherjee2020uncertainty} are used as baselines in our paper.

}


\noindent\textbf{Light-weight tuning.} Standard fine-tuning methods tune all trainable model parameters for every task. Recent efforts have focused on lightweight tuning of large PLMs by updating a small set of parameters while keeping most of parameters in PLMs frozen, including prefix tuning~\citep{prefix}, prompt token tuning~\citep{prompt_tuning} and Adapter tuning~\citep{houlsby2019parameter,pfeiffer2020AdapterHub}. All of the above works focus on fully supervised settings with thousands of labeled examples using classic fine-tuning. In contrast, we focus on few-shot learning settings leveraging prompts for model tuning, where we make several observations regarding the design and placement of adapters in few-shot settings in contrast to its resource-rich counterpart. {Some recent works~\citep{beck2021adapterhub, zhong2021useradapter} pre-train adapters with full supervision with thousands of labeled examples from source tasks for few-shot target adaptation. Different from this, we explore adapter tuning for single-task few-shot learning without any auxiliary supervision labels.}

\section{ Methodology}
\subsection{Overview}
\begin{figure}[hbt!]
\centering
\vspace{-0.in}
\includegraphics[width=0.8\linewidth]{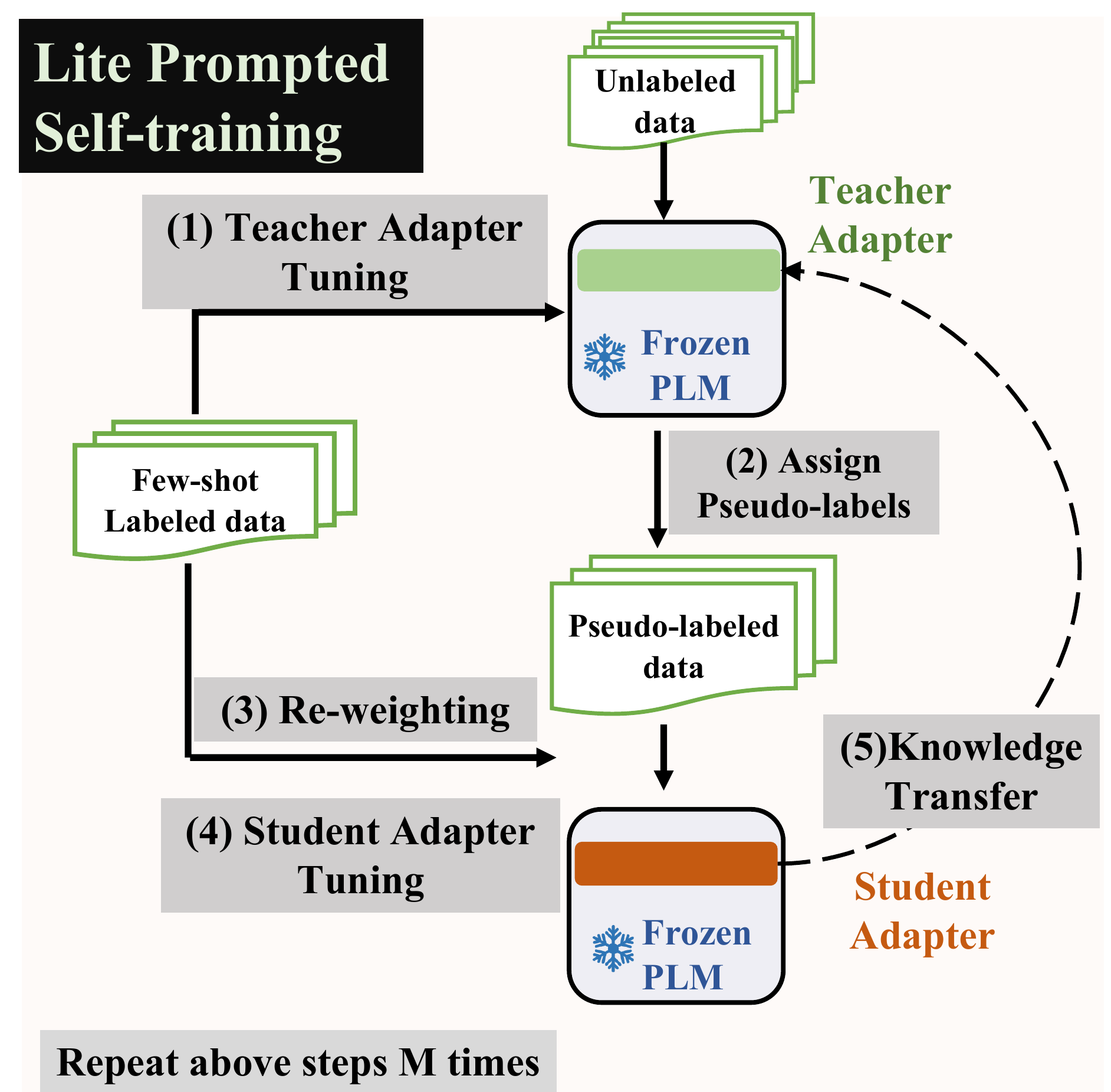}
\caption{Lite prompted self-training on unlabeled data with prompts and adapters make parameter-efficient few-shot learners with {\tt LiST}.}\label{fig:framework}
\end{figure}

We adopt a PLM (e.g., RoBERTa~\citep{RoBERTa}) as the {\em shared encoder} for both the student and teacher for self-training. The shared PLM encoder is frozen and not updated during training. We introduce {\em tunable adapter parameters} in both teacher and student (discussed in Section~\ref{subsec:prompt-adapter}) that are iteratively tuned during self-training. Refer to Figure~\ref{fig:framework} for steps in the following discussion.

We first use prompt-based fine-tuning to update the teacher adapter ({\em Step 1}) with few-shot labeled examples and leverage the teacher model to assign pseudo-prompt labels ({\em Step 2}) on unlabeled data $\mathcal{D}^u$. The teacher is often uncertain in few-shot learning and produces noisy pseudo-labels. Therefore, we adopt meta-learning (discussed in Section~\ref{subsec:reweight}) to re-weight the noisy pseudo-labeled samples ({\em Step 3}). The re-weighted data is used to train the student adapter ({\em Step 4}). Since adapter training with noisy pseudo labels is quite unstable, we introduce knowledge distillation warmup (discussed in Section~\ref{subsec:kd}).  Finally, we assign the trained student adapter to be the new teacher adapter ({\em Step 5}). Following \emph{true} few-shot learning settings, we do not use any held-out development or validation set. 
Therefore, we repeat the above steps for a pre-defined number of times ($M=6$). The overall training procedure is summarized in Algorithm~\ref{alg} (Appendix~\ref{sec:alg}). Throughout the training, we keep the shared student and teacher encoder parameters frozen and update the corresponding adapter parameters along with their language model heads.

\subsection{Lightweight Prompt Adapter Tuning}
\label{subsec:prompt-adapter}

The predominant methodology for task adaptation is to tune all of the trainable parameters of the PLMs for every task. This raises significant resource challenges both during training and deployment. A recent study~\citep{aghajanyan2020intrinsic} show that PLMs have a low instrinsic dimension that can match the performance of the full parameter space. 
To adapt PLMs for downstream tasks with a small number of parameters, adapters~\citep{houlsby2019parameter} have recently been introduced as an alternative approach for lightweight tuning. Consider the following scenario for demonstration, where we want to use RoBERTa-large with $\mathcal{M}=355M$ parameters as the PLM for $\mathcal{T}=100$ tasks. Full fine-tuning for this scenario requires updating and storing $\mathcal{M}\times\mathcal{T}=35.5B$ parameters. Now, consider fine-tuning with {\sysname} that requires $\mathcal{A}=14M$ (tunable) adapter parameters for every task while keeping the PLM fixed. This results in overall $\mathcal{M}+\mathcal{A}\times\mathcal{T}=1.8B$ parameters, thereby, reducing the overall storage cost by 20x. Adapters have been shown to match the PLM performance in {fully supervised settings with thousands of training labels in classic fine-tuning}. {In contrast, this is the first work to study the role of adapters in few-shot prompt-based FN.} We explore different design and placement choices of adapters in few-shot settings and investigate the performance gap with fully supervised as well as fully tunable parameter space. 

\begin{figure}[hbt!]
 	\centering
   	\includegraphics[width=0.7\linewidth]{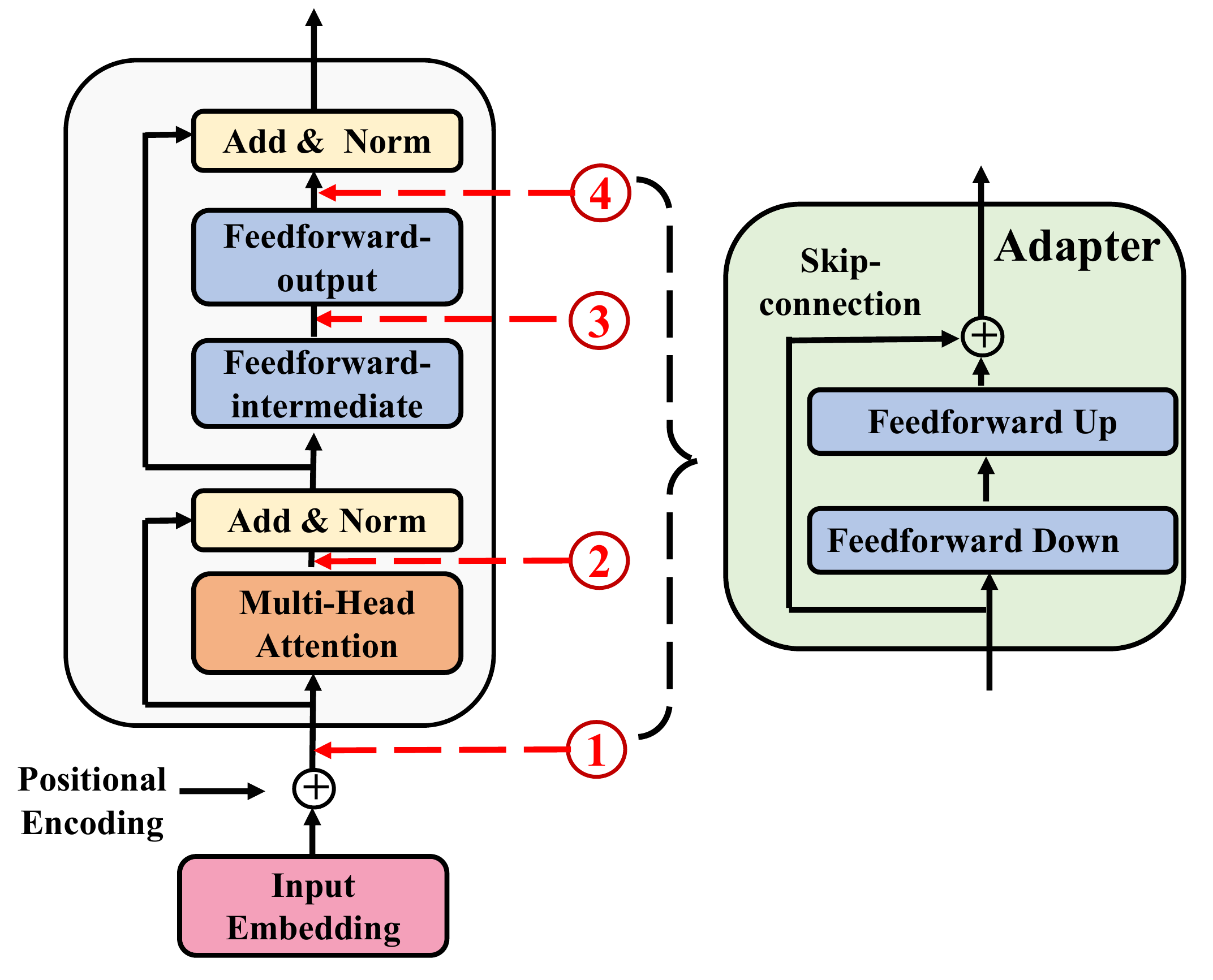}\vspace{-0.5em}
 	  	\caption{{\sysname} explores several adapter placement choices (numbered positions in left) in standard Transformer architecture, with adapter design shown in right.}
 	 	 	\label{fig:adapter}
\end{figure}

The adapter tuning strategy judiciously introduces new parameters into the original PLMs.  In contrast to standard prompt-based FN  that updates all the PLM parameters $\Theta_{\mathrm{PLM}}$, prompt-adapter tuning only updates the newly introduced adapter parameters as well as the (masked) language model head of the PLM (jointly denoted as $\psi$),  while keeping the remaining parameters of the original network frozen. {The adapter used in {\sysname} consists of two fully connected layers as shown in Figure~\ref{fig:adapter}, where a feedforward layer down projects input representations to a low dimensional space $d$ (referred as the bottleneck dimension), and another feedforward layer up projects the low-dimensional features back to the original dimension. However, these newly-inserted parameters can cause divergence resulting in up to $20\%$ performance degradation in few-shot settings (discussed in Section~\ref{subsec:adapter}). To handle this issue, we adopt a skip-connection design where the adapter parameters are initialized with zero-mean small Gaussian noise.}


\vspace{-0.05in}

\noindent\textbf{Adapter placement.} Prior works on lightweight adaptation tune bias~\citep{cai2020tinytl} or embeddings~\citep{prompt_tuning} of Transformers in fully-supervised settings for improving parameter-efficiency with minimal performance loss. However, for few-shot settings, we note that adapter placement is critical to bridge the performance gap with that of a fully tunable model and the choices of tuning bias or embedding can result in upto 10\% performance degradation (discussed in Section~\ref{subsec:adapter}). To this end, we explore several choices of adapter placement (refer to Figure~\ref{fig:adapter}) corresponding to the most important transformer modules, namely, embedding, intermediate feedforward, output feedforward and attention module in {\em every layer} of the Transformer. Based on empirical experiments (refer to Section~\ref{subsec:adapter}) across six diverse NLU tasks, we observe the feedforward output and attention modules to be the most important components for parameter-efficient adaption in few-shot settings.


Formally, consider $\widetilde{\mathcal{D}}^{Train}_K=\{\widetilde{x}^l, \widetilde{y}^l\}$ to be the few-shot labeled data and $\widetilde{\mathcal{D}}^U=\{\widetilde{x}_u\}$ to be the unlabeled data, where we transform the input sequences $x$ to cloze-style input $\widetilde{x}$ containing a single mask  following the prompting strategy outlined in Section~\ref{subsec:primer}. We use the same pattern templates and verbalizers (output mapping from the task-specific labels $\mathcal{Y}$ to single tokens in the vocabulary $\mathcal{V}$) from traditional prompt-based FN  works~\citep{gao2021making}. Given the above adapter design and placement of choice with parameters $\psi$, a dataset  $\widetilde{\mathcal{D}}_K^{Train}$ with shots $K$, a PLM encoder $enc$ with parameters $\Theta_{\mathrm{PLM}}$, where $\Theta_{\mathrm{PLM}} \gg \psi$, we want to perform the following optimization for efficient model adaptation:
\begin{equation}
\small
\psi \leftarrow \argmin_{\psi}\  \mathcal{L} (\widetilde{\mathcal{D}}_K^{Train}; \Theta_{\mathrm{PLM}}, \psi)\vspace{-0.1in}
\end{equation}
\vspace{-0.15in}
\subsection{Re-weighting Noisy Prompt Labels}\
\label{subsec:reweight}
\vspace{-0.1in}



Consider $\{\hat{y}^{(t)}_{n}\}_{n=1}^N$ to be the pseudo prompt-labels (for the masked tokens in $\widetilde{x}^u_{n} \in \widetilde{X}$) from the teacher $(\Theta_{\mathrm{PLM}}, \hat{\psi}_{\mathrm{tea}})$ in the $t$-th iteration where $N$ is the number of unlabeled instances and $\hat{\psi}_{\mathrm{tea}}$ represent the teacher adapter parameters. In self-training, the student model is trained to mimic the teacher predictions on the transfer set. Consider $\mathcal{L}(\hat{y}^{(t)}_{n}, \mathrm{enc}(\widetilde{x}^u_{n}; \Theta_{\mathrm{PLM}}, \psi^{(t)}_{\mathrm{stu}}))$ to be the loss of the student model with parameters $(\Theta_{\mathrm{PLM}}, \psi^{(t)}_{\mathrm{stu}})$ on the pseudo-labeled data in the $t$-th iteration, where $\Theta_{\mathrm{PLM}}$ and $\psi_{\mathrm{stu}}$ represent the PLM and the student adapter parameters respectively.
In order to reduce error propagation from noisy pseudo-labels, we leverage meta-learning to re-weight them based on the student model loss on the validation set as our meta-objective.  The intuition of meta re-weighting is to measure the impact or weight of a pseudo-labeled example given by its performance on the {\bf validation set}. Since we do not have access to a separate validation set in the spirit of true few-shot learning, we leverage the labeled training set $\widetilde{\mathcal{D}}^{Train}_K$ judiciously for re-weighting. To this end, we leverage the idea of weight perturbation~\citep{ren2018learning} to set the weight of pseudo-labeled example $(\widetilde{x}^u_{i}, \hat{y}^{(t)}_{i})$ to $\epsilon^{(t)}_{i}$ at iteration $t$ as:
\begin{equation}
\small
    \mathcal{L}^{(t)}_{r} (\epsilon, \psi) = \frac{\sum_{i=1}^{N} [\epsilon^{(t)}_{i} \cdot \mathcal{L}(\hat{y}^{(t)}_{i}, \mathrm{enc}(\widetilde{x}^u_{i}; \Theta_{\mathrm{PLM}},  \hat{\psi}^{(t-1)}_{\mathrm{stu}}))] }{N}.
    \vspace{-0.5em}
\end{equation}
\begin{equation}
\small
    \hat{\psi}^{(t)}_{\mathrm{stu}} (\epsilon) = \hat{\psi}^{(t-1)}_{\mathrm{stu}} - \alpha \triangledown \mathcal{L}^{(t)}_{r} (\epsilon, \psi) .
    \vspace{-0.5em}
\end{equation}
{where $\alpha$ is the step size. Weight perturbation is used to discover data points that are most important to improve performance on the validation set. Optimal value for the perturbation $\epsilon^{(t)*}_{i}$ can be obtained via minimizing student model loss on the validation set at iteration $t$ as:}
\begin{equation}
\small
\epsilon^{(t)*}_{i} = \argmin_{\epsilon_{i}} \frac{  \sum_{i=1}^{|\widetilde{\mathcal{D}}^{Train}_K|} \mathcal{L}({y}_{i}, \mathrm{enc}(x_{i}; \Theta_{\mathrm{PLM}}, \hat{\psi}^{(t)}_{stu}(\epsilon_{i}))}{|\widetilde{\mathcal{D}}^{Train}_K|}
\end{equation}
 To obtain a cheap estimate of the meta-weight at step $t$, we take a single gradient descent step on a mini-batch $\widetilde{\mathcal{D}}^{(t)} \in \widetilde{\mathcal{D}}^{Train}_K$ as:
\begin{equation}
\small
    u_{i}^{(t)} = -\frac{\partial}{\partial{\epsilon_{i}}}  \bigg ( \frac{  \sum_{i=1}^{|\widetilde{\mathcal{D}}^{(t)}|}  \mathcal{L}(y_{i}, \mathrm{enc}(\widetilde{x}_{i}; \Theta_{\mathrm{PLM}},  \hat{\psi}^{(t)}_{stu} (\epsilon)))}{{|\widetilde{\mathcal{D}}^{(t)}|}} \bigg) 
\label{eq:grad}
\end{equation}
{The weight  $w_{i}^{(t)}$  of $(\widetilde{x}^u_{i}, \hat{y}^{(t)}_{i})$ at iteration $t$ is set to be proportional to the negative gradient $u_{i}^{(t)}$ to reflect the importance of pseudo-labeled samples. Samples with negative weights are filtered out since they could potentially degrade the student performance. Finally, we update student adapter parameters $\psi_{\mathrm{stu}}$ while accounting for re-weighting as:}
\begin{equation}
\small
\label{Eq:meta_update}
    \mathcal{L}^{(t)} =  \frac{1}{N}  \sum_{i=1}^{N} [w_{i}^{(t)} \cdot \mathcal{L}(\hat{y}^{(t)}_{i}, \mathrm{enc}(\widetilde{x}^u_{i}; \Theta_{\mathrm{PLM}}, \hat{\psi}^{(t-1)}_{\mathrm{stu}}))] \big ). 
\end{equation}



 
\subsubsection{Knowledge Distillation For Student Warmup}

\label{subsec:kd}
Meta re-weighting leverages gradient as a proxy to estimate the weight of noisy pseudo labels. However, the gradients of adapter parameters $\psi$ are not stable in the early stages of training due to random initialization and noises in pseudo labels. This instability issue is further exacerbated with adapter tuning that usually requires a larger learning rate~\citep{pfeiffer2020AdapterHub}. Therefore, to stabilize adapter tuning, we propose a warmup training stage via knowledge distillation~\citep{hinton2015distilling} to first tune adapter parameters via knowledge distillation loss for $T_{warm}$ steps and then we continue self-training with re-weighted updates via Eq.~\ref{Eq:meta_update}. Since the re-weighting procedure has access to our training labels, we do not use labeled data in knowledge distillation while using only the unsupervised consistency loss between teacher model $(\Theta_{\mathrm{PLM}}, \hat{\psi}_{\mathrm{tea}})$ and student model $(\Theta_{\mathrm{PLM}}, \hat{\psi}_{\mathrm{stu}})$ on unlabeled data as.
\begin{equation}
\small
\label{Eq:kd_warmup}
   \argmin_{\hat{\psi}_{stu}} \infdiv{f(\widetilde{x}^u; \Theta_{\mathrm{PLM}}, \hat{\psi}_{\mathrm{tea}})}{f(\widetilde{x}^u; \Theta_{\mathrm{PLM}}, \hat{\psi}_{\mathrm{stu}}) }. 
\end{equation}
We further validate the effectiveness of knowledge distillation for warmup with ablation analysis. 

\subsubsection{Student Adapter Re-initialization}

A typical challenge in few-shot settings is the lack of a separate validation set. In the spirit of {\em true} few-shot learning, we use only the available few-shot labeled examples $\widetilde{\mathcal{D}}^{Train}_K$ as the validation set for meta-learning of the student model. This poses an interesting challenge of preventing label leakage.  To address this issue, we {\em re-initialize the student adapter parameters} every time at the start of each self-training iteration to mitigate interference with labeled data. Note that the student and teacher model share the encoder parameters $\Theta_{\mathrm{PLM}}$ that are always kept frozen and not updated during training.

\section{Experiments}
\label{sec:exp}
\subsection{Experimental Setup}
\label{subsec:setup}
\noindent\textbf{Dataset.} We perform large-scale experiments with six natural language understanding tasks as summarized in Table~\ref{tab:datasets}. We use four tasks from GLUE~\citep{wang2019glue}, including MNLI~\citep{williams2018broad-mnli} for natural language inference, RTE~\citep{dagan2005pascal-rte1,bar2006second,giampiccolo2007third-rte3,bentivogli2009fifth-rte4} for textual entailment, QQP\footnote{\url{https://www.quora.com/q/quoradata/}} for semantic equivalence and SST-2~\citep{socher2013recursive-sst-2} for sentiment classification. The results are reported on their development set following~\citep{zhang2020revisiting}.  MPQA~\citep{wiebe2005annotating} and Subj~\citep{pang2004sentimental-subj} are used for polarity and subjectivity detection, where we follow~\cite{gao2021making} to keep $2,000$ examples for testing and use remaining examples for semi-supervised learning. 

For each dataset, we randomly sample $|\mathcal{K}| \in \{10, 20, 30\}$ manually labeled samples from the training data, and add the remaining to the unlabeled set while ignoring their labels -- following standard setups for semi-supervised learning. {We repeatedly sample $K$ labeled instances five times, run each model with $5$ different seeds and report average performance with standard deviation across the runs.} For the average accuracy over 6 tasks, we did not include standard deviation across tasks.  Furthermore, for every split and shot, we sample the labeled data such that $\mathcal{D}_{10}^{Train} \subset \mathcal{D}_{20}^{Train} \subset \mathcal{D}_{30}^{Train}$ to evaluate the impact of incremental sample injection. 

Following \textit{true few-shot learning} setting~\citep{perez2021true}, we do not use additional \textit{development set} beyond $|\mathcal{K}|$ labeled samples for any hyper-parameter tuning or early stopping. The performance of each model is reported after fixed training epochs (see  Appendix for details).


\begin{table*}[h]
\centering
\resizebox{\textwidth}{!}{
\begin{tabular}{clllllllll}
\toprule
\textbf{Labels} & \textbf{Models} & {\textbf{Avg}} & \textbf{\#Tunable} &\multicolumn{1}{l}{\textbf{MNLI (m/mm)}}  & \multicolumn{1}{l}{\textbf{RTE}}   & \multicolumn{1}{l}{\textbf{QQP}} & \multicolumn{1}{l}{\textbf{SST-2}} & \multicolumn{1}{l}{\textbf{Subj}} & \multicolumn{1}{l}{\textbf{MPQA}} \\
&&&{\bf Params}&(acc) &(acc) &(acc) & (acc)& (acc)&(acc)\\
\midrule

\multirow{2}{*}{\thead{$|\mathcal{K}| = 30$ }}

& Classic FN &60.9 &355M &38.0 {\tiny(1.7)} / 39.0 \tiny{(3.1)} & 51.4 {\tiny(3.7)} &64.3 {\tiny(8.1)}  & 65.0 {\tiny(11.5)} & 90.2 {\tiny(2.2)} &56.1 {\tiny(5.3)}\\
& Prompt FN & 77.6 &355M &62.8 {\tiny(2.6)} / 64.1 {\tiny(3.3)} & 66.1 {\tiny(2.2)} &  71.1 {\tiny(1.5)} & 91.5 {\tiny(1.0)} & 91.0 {\tiny(0.5)} &82.7 ({\tiny3.8)} \\

\midrule
\multirow{4}{*}{\thead{$|\mathcal{K}| = 30$ \\  +Unlabeled Data}}

&UST& 65.8 & 355M & 40.5 {\tiny (3.3)} / 41.5 {\tiny (2.9)} & 53.4 {\tiny (1.7)} & 61.8 {\tiny (4.3)} & 76.2 {\tiny (11.4)} & 91.5 {\tiny (2.1)} & 70.9 {\tiny (6.2)} \\
&MetaST& 62.6 & 355M &39.4 {\tiny(3.9)} / 40.5 {\tiny(4.4)} & 52.9 {\tiny(2.0)} & 65.7 {\tiny(6.2)} 
 & 65.3 {\tiny(15.2)}& 91.4 {\tiny(2.3)} & 60.5 {\tiny(3.6)}\\

&iPET & 75.5 & 355M&61.0 {\tiny(5.8)} / 61.8 {\tiny(4.7)} & 54.7 {\tiny(2.8)} & 67.3 {\tiny(4.1)}  & \textbf{93.8 {\tiny(0.6)}} & 92.6 {\tiny(1.5)}& 83.1 {\tiny(4.8)}\\
& PromptST & 77.2 & 14M &  61.8 {\tiny(1.9)} / 63.1 {\tiny(2.9)} & 66.2 {\tiny(5.1)} & 71.4 {\tiny(2.1)} & 91.1 {\tiny(1.4)} & 90.3 {\tiny(1.5)}
 & 81.8 {\tiny(2.5)}\\


 &{\sysname}& \textbf{82.0} &14M& \textbf{73.5 {\tiny(2.8)} / 75.0 {\tiny(3.7)}} &\textbf{71.0 {\tiny(2.4)}}&\textbf{75.2 {\tiny(0.9)} }& 92.8 {\tiny(0.9)} &\textbf{93.5 {\tiny(2.2)}} & \textbf{85.2 {\tiny(2.1)}} \\
 \midrule
Supervision with & Classic FN & 90.9 & 355M & 89.6 / 89.5 &83.0& 91.8 &95.2&97.2&88.8 \\
\# Full Train & Prompt FN & 92.0 & 355M&89.3 / 88.8 & 88.4 & 92.1  & 95.9& 97.1 & 89.3

 \\
 \midrule
 
\end{tabular}
}\vspace{-0.in}
\caption{Performance comparison of different tuning strategies on different NLU tasks with RoBERTa-large as the encoder with standard deviation in parantheses. UST, MetaST, PromptST and iPET are semi-supervised methods using unlabeled data, whereas Classic and Prompt FN only use labeled data.}
\label{tab:main_results}
\vspace{-0.in}
\end{table*}
\noindent\textbf{Baselines.} In addition to classic-tuning (Classic FN), we adopt prompt-based fine-tuning (Prompt FN) from~\cite{gao2021making} as labeled-only baselines. We also adopt several state-of-the-art semi-supervised baselines including UST~\citep{mukherjee2020uncertainty}, MetaST~\citep{wang2021meta} and iPET~\citep{schick2021exploiting}. UST and MetaST are two self-training methods which are based on classic fine-tuning strategies.  iPET is a semi-supervised method leveraging prompt-based fine-tuning and prompt ensembles to obtain state-of-the-art performance. {While iPET ensembles multiple fully-tuned models, we develop a lite self-training framework to achieve both data and parameter efficiency.} {As the strongest semi-supervised baseline, we implement a new method {\em PromptST based on self-training using prompts and adapters} (as a subset of the methods used in {\sysname}), but without any re-weighting, or KD warmup that are additionally used in {\sysname}. {The methods Prompt FN, PromptST and {\sysname} adopt same prompts and label words as in \cite{gao2021making}.} We implement our framework in Pytorch and use Tesla V100 gpus for experiments. Prompts used in experiments and hyper-parameter configurations are presented in Appendix.}


\vspace{-0.in}
\subsection{Key Results}
Table~\ref{tab:main_results} shows the performance comparison among different models with $|\mathcal{K}| = 30$ labeled examples with fixing RoBERTa-large as the encoder.    Fully-supervised RoBERTa-large trained on thousands of labeled examples provides the ceiling performance for the few-shot setting. We observe {\sysname} to significantly outperform other state-of-the-art baselines along with $96\%$ reduction in tunable parameters, achieving both labeled data- and parameter-efficiency. {More specifically, {\sysname} improves over Classic FN, Prompt FN, iPET and PromptST by $34.6\%$, $5.7\%$, $8.6\%$ and $6.2\%$ respectively in terms of average performance on six tasks.} This demonstrates the impact of self-training with unlabeled data and prompt-based FN. Additionally, iPET and {\sysname} both leverage prompt-based FN to significantly improve over UST and MetaST that use classic fine-tuning strategies, confirming the effectiveness of prompt-based FN in the low data regime. {iPET ensembles multiple prompts with diverse qualities and under-performs Prompt FN on average in our few-shot setting without using any development set.  }

\begin{figure}[hbt!]
\centering
\subfloat[MNLI]{
\begin{minipage}{0.8\linewidth}
\includegraphics[width=1.0\linewidth]{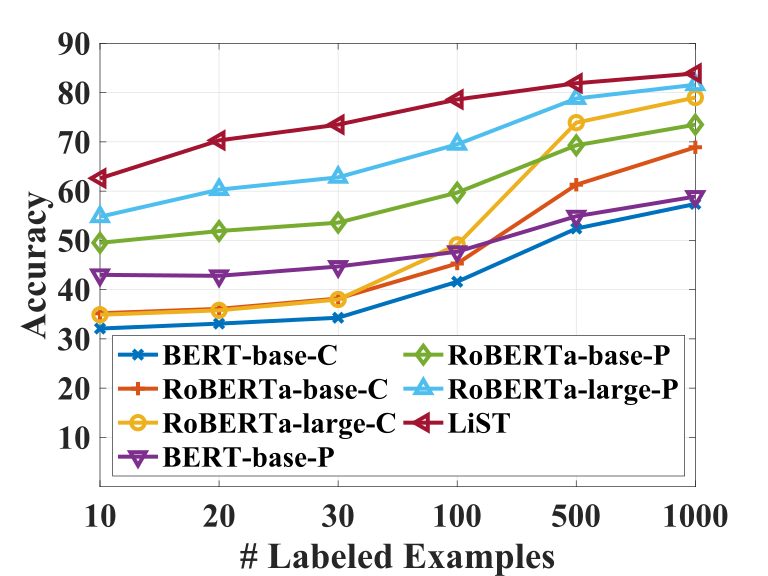}\vspace{-0.15in}
\label{subfig:mnli}
\end{minipage}}

\subfloat[RTE]{
\begin{minipage}{0.8\linewidth}
\includegraphics[width=1.0\linewidth]{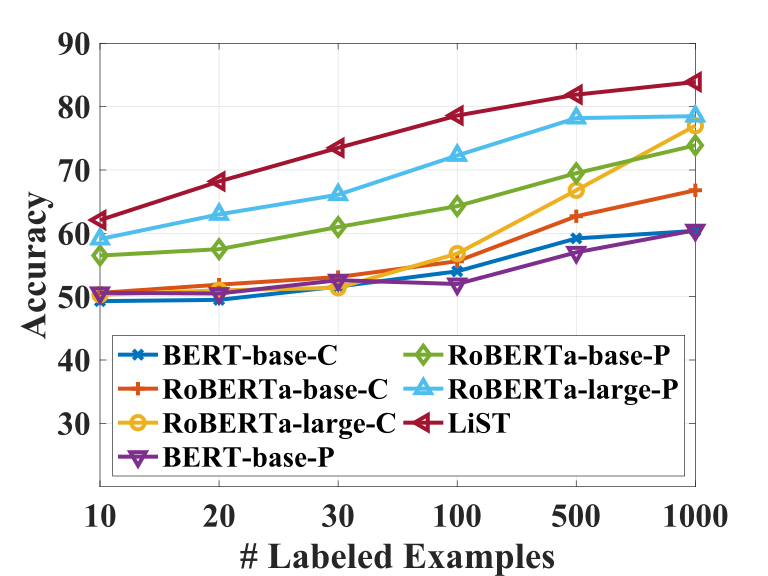}\vspace{-0.15in}
\label{subfig:rte}
\end{minipage}}

 \caption{Performance comparison of Classic-tuning (denoted as ``C") and prompt-based fine-tuning (denoted as ``P") with {\sysname} on MNLI and RTE using language model encoders of different sizes. }\label{fig:main_fig}
\end{figure}

\begin{table*}[h]

\centering
\resizebox{\textwidth}{!}{
\begin{tabular}{cllllllll}
\toprule
\textbf{Labels} & \textbf{Fine-tuning Method} & \textbf{Avg} & \multicolumn{1}{l}{\textbf{MNLI (m/mm)}}  & \multicolumn{1}{l}{\textbf{RTE}}   & \multicolumn{1}{l}{\textbf{QQP}} & \multicolumn{1}{l}{\textbf{SST-2}} & \multicolumn{1}{l}{\textbf{Subj}} & \multicolumn{1}{l}{\textbf{MPQA}}  \\
&&&(acc) &(acc) &(acc) & (acc)& (acc)&(acc)\\
\midrule

\multirow{3}{*}{$|\mathcal{K}| = 10$}
& GPT-3 In-context  & 61.5 & 36.4 {\tiny(0.8)} / 36.7 {\tiny(1.3)} &  53.2 {\tiny(1.8)} & 61.8 {\tiny(3.0)} & 86.6 {\tiny(7.4)} &  61.0 {\tiny(11.2)} & 66.7 	{\tiny(9.5)}\\
& Prompt-based FN  &69.3 &  54.8 {\tiny(3.7)} / 55.6 {\tiny(4.6)} & 60.0 {\tiny(4.4)} & 58.7 {\tiny(4.6)} & 89.5 {\tiny(1.7)} & 84.5 {\tiny(8.6)} & 67.8 {\tiny(6.9)}\\
&{\sysname} & 72.8 &62.6 {\tiny(6.6)} / 63.3 {\tiny(7.7)} & 61.2 {\tiny(4.9)} & 60.4 {\tiny(7.0)} & 91.1 {\tiny(1.2)} & 91.0 {\tiny(1.6)}& 70.3 {\tiny(10.6)}\\
\midrule

\multirow{3}{*}{$|\mathcal{K}| = 20$}
& GPT-3 In-context & 57.4 &   38.0 {\tiny(2.0)} / 38.4 {\tiny(2.8)} & 54.5 {\tiny(1.5)} & 64.2 {\tiny(1.6)}  & 79.1 {\tiny(2.3)} & 51.2 {\tiny(1.7)} & 72.4 {\tiny(8.5)} \\
& Prompt-based FN  &75.4 &  60.3 {\tiny(2.0)} / 61.6 {\tiny(2.7)} & 64.3 {\tiny(2.4)} & 67.8 {\tiny(4.2)}& 90.6 {\tiny(1.8)} &  88.3 {\tiny(2.2)} & 80.6 {\tiny(7.5)} \\
& {\sysname} &  79.5 & 68.9 {\tiny(3.1)} / 70.4 {\tiny(3.3)} & 69.0 {\tiny(3.5)} & 72.3 {\tiny(3.7)} & 92.3 {\tiny(1.2)} &  91.5 {\tiny(1.3)} & 82.2 {\tiny(5.1)}\\
\midrule

\multirow{3}{*}{$|\mathcal{K}| = 30$}

&  GPT-3 In-context & 61.5 & 37.9 {\tiny(2.2)} / 38.5 {\tiny(2.9)} & 53.4 {\tiny(2.2)} &65.0 {\tiny(1.7)}  & 79.7 {\tiny(7.1)} & 57.7 {\tiny(6.4)} &74.8 {\tiny(6.9)} \\

& Prompt-based FN  & 77.6 &  62.8 {\tiny(2.6)} / 64.1 {\tiny(3.3)} & 66.1 {\tiny(2.2)} &  71.1 {\tiny(1.5)} & 91.5 {\tiny(1.0)}& 91.0 {\tiny (0.5)}&82.7 {\tiny(3.8)} \\

& {\sysname}& {82.0} & {73.5 {\tiny(2.8)} / 75.0 {\tiny(3.7)}} &{71.0 {\tiny(2.4)}}&{75.2 {\tiny(0.9)} }& 92.8 {\tiny(0.9)} &{93.5 {\tiny(2.2)}} & {85.2 {\tiny(2.1)}}  \\

\midrule
\end{tabular}
}
\caption{Average performance and standard deviation of GPT-3 ($175B$ params) in-context learning, Prompt-based FN and {\sysname} methods using Roberta-Large ($355M$ params) encoder with varying number of training labels
$|\mathcal{K}|$. {\sysname} updates $14M$ params in contrast to Prompt-based FN with full model tuning.}
\label{tab_app:gpt-3}
\end{table*}

Figure~\ref{fig:main_fig} compares the performance of tuning methods with varying number of training labels and encoders of different sizes. We observe that large models are more data-efficient compared to smaller models. However, large fully-tunable models are expensive to use in practise. We observe that {\sysname} with small number of tunable parameters consistently outperforms fully-tunable classic and prompt-based FN strategies in all labeled data settings, demonstrating both data and parameter efficiency. {Additional results with different backbone encoders and varying number of shots and fine-tuning strategies are presented in the Appendix in Tables~\ref{tab_app:var_encoder_30}, ~\ref{tab_app:var_encoder_20}, ~\ref{tab_app:var_encoder_10} and~\ref{tab:different_fn} that demonstrate similar trends as we observe in Table~\ref{tab:main_results} and Figure~\ref{fig:main_fig}}. 

\noindent\textbf{Comparison with GPT-3 in-context Learning.}
We perform a comparison between GPT-3 in-context learning, RoBERTa-large Prompt-based fine-tuning and LiST methods with varying number of training labels in Table~\ref{tab_app:gpt-3}. For a fair comparison, the prompt and label words are same for the three approaches. We observe that {\sysname} outperforms GPT-3 In-context learning and Prompt-based FN consistently with different number of labels.

\subsection{Adapter Analysis}
\label{subsec:adapter}
\vspace{-0.05in}

In this section, we explore adapter design choices for prompt-based FN with RoBERTa-large as encoder {\em using only few-shot labeled data}. 

\begin{table}
\centering
\resizebox{0.8\linewidth}{!}{
\begin{tabular}{cllll}
\toprule
 \textbf{Tuning} & \textbf{\#Params} &  \textbf{Avg} & \textbf{Diff} \\

\midrule










 Full & 355M &77.6 & ---\\
\cdashline{2-5}
 Embedding &53M &67.0 & -10.7 \\
 Attention &101M & 77.0 & -0.6\\
 FF-output & 102M&\textbf{77.6} & +0.0

\\
 FF-intermediate  &102M & 75.9 & -1.7 

\\

\midrule

\end{tabular}}
\caption{Average accuracy on tuning different modules of RoBERTa-large with $|\mathcal{K}|=30$ labels on \textbf{six tasks}. {Diff shows performance change relative to Full tuning.} }
\label{tab:adapter_pilot}\vspace{-0.1in}
\end{table}

\noindent\textbf{Where to insert an adapter in Transformers?} In order to answer this question, we conduct an experiment to study the role of various Transformer modules in few-shot prompt-based FN. To this end, we tune a given module along with the language model head while keeping all other parameters frozen. Table~\ref{tab:adapter_pilot} shows the performance comparison of tuning specific modules on six tasks with varying number of labeled examples. The main modules of RoBERTa include \textit{Embedding}, \textit{Attention}, \textit{Feedforward Output} and \textit{Feedforward Intermediate} layers. We observe that tuning only the \textit{Feedforward Output} or the \textit{Attention} module delivers the best performance across most tasks with few-shot labels. Correspondingly, this motivated us to insert our adapter parameters into these two modules. {More detailed results are presented in Appendix Table~\ref{tab_app:adapter_pilot}}. 

\begin{table}

\centering
\resizebox{\linewidth}{!}{
\begin{tabular}{llcl}
\toprule
 \textbf{Tuning} & \textbf{\#Params}& \textbf{Avg} \\
\midrule

 Head-only &1M&66.9 \\
 Bias-only~\citep{cai2020tinytl}& 1M &68.3  \\
 
 Prompt-tuning~\citep{lester2021power} &1M& 56.4\\

{{\sysname} Adapter (2) }& 1M & 72.7\\ \hdashline
 
  Houlsby Adapter~\citep{houlsby2019parameter} &14M& 57.9\\
 {\sysname} Adapter (128)  & 14M & \textbf{77.7}\\ 
 \midrule
  Full tuning & 355M& 77.6 \\

\midrule
 
\end{tabular}
}
\caption{Average accuracy of several lightweight parameter-efficient tuning strategies with $|\mathcal{K}| = 30$ labels without unlabeled data on \textbf{six tasks} along with the number ($\#$) of tunable parameters. Each task is run with 5 different seeds. {\sysname} Adapter performance with different bottleneck dimension $d$ of its adapters is shown in parantheses. }
\vspace{-0.1in}
\label{tab:adapter}
\end{table}

\noindent\textbf{Comparison with other lightweight parameter efficient model tuning strategies.} To validate the effectiveness of {\sysname} adapters, we compare it against several baselines in Table~\ref{tab:adapter}. {For a fair comparison, we present two variations of our {\sysname} adapters with bottleneck dimensions $d$= $\{2, 128\}$ corresponding to $1M$ and $14M$ parameters to match other adapter capacities; all the approaches in Table~\ref{tab:adapter} are {\em trained with 30 labels only without unlabeled data} for a fair comparison.  (1) Bias-only is a simple but effective lightweight method, which tunes bias terms of PLMs while keeping other parameters frozen. (2) Tuning head layers is widely used as a strong baseline for lightweight studies~\citep{houlsby2019parameter}, where we tune last two layers including language model head while freezing other parameters. (3) prompt-tuning is a lightweight method which only updates task prompt embedding while keeping entire model frozen. (4) Houlsby Adapter tunes inserted adapter parameters keeping the encoder frozen by adopting classic tuning strategy.}
Besides these lightweight methods, we also present a performance comparison with full model tuning as a strong baseline. {More detailed results are presented in Appendix in Tables~\ref{tab_app:adapter} and ~\ref{tab_app:adapter_var_shots} that demonstrate similar trends.}

{Table~\ref{tab:adapter} shows that {\sysname} is able to match the performance of full model prompt-based FN  with bottleneck dimension $d=128$ and outperforms all other baselines with similar capacities. While lightweight model tuning choices like tuning the bias or inserting adapters into classic tuning models are shown to be effective in fully-supervised settings~\citep{cai2020tinytl, houlsby2019parameter}, we observe them to under-perform for few-shot learning. 
{We observe that simpler tuning choices like Head-only and Bias-only results in upto $10\%$ performance degradation. Houlsby adapter and Prompt-only results in upto $20\%$ performance degradation. In constrast, {\sysname} adapter is able to match the performance of full tuning in few-shot setting, demonstrating the importance of adapter placement choices and parameter initialization.}
}

\vspace{-0.1in}
\subsection{Ablation Analysis}

Table~\ref{tab:ablation_model} demonstrates the impact of different components and design choices of {\sysname}.

\noindent$\bullet$ \textbf{Adapter training stability.} Training with very few labels and noisy pseudo labeled data results in instability for adapter tuning. To demonstrate training stability, we include the average accuracy and standard deviation across several runs and splits as metrics. We observe that hard pseudo-labels hurt the model performance compared to soft pseudo-labels and exacerbate the instability issue. This is in contrast to observations from classic fine-tuning~\citep{wang2021meta}. A potential reason could be that the well pre-trained language model head for prompt-based FN is able to capture better associations among different prompt labels.

\begin{table}
\vspace{-0.in}
\resizebox{\linewidth}{!}{
\begin{tabular}{lcclllll}
\toprule
 \multirow{2}{*}{\textbf{Method}} &  \multirow{2}{*}{\textbf{Avg Acc}} & \multirow{2}{*}{\textbf{Avg Std}} & 
\multicolumn{3}{c}{\textbf{Datasets}}\\
\cmidrule{4-6}
 & && MNLI (m/mm) & RTE   \\
\midrule
\midrule
{\sysname} ($14M$)  & 72.6 &2.8 & 73.5 {\tiny(2.8)} / 75.0 {\tiny(3.7)} & 71.0 {\tiny(2.4)} \\

{w/o re-init}	& {68.3} & {4.2} &{ 66.7 {\tiny(2.8)} / 68.3 {\tiny(4.3)}} &	{69.0 {\tiny(4.9)}} \\
 w/o KD Warmup & 68.8 & 8.8 &67.9 {\tiny(12.9)} / 69.0 {\tiny(13.1)} & 69.2 {\tiny(4.5)} 
 \\
w/o Re-weighting &  71.6 & 4.0&72.9 {\tiny(3.4)} / 74.2 {\tiny(4.5)} & 69.7 {\tiny(4.1)} 
 \\
 w/ Hard Pseudo-Labels & 70.9 & 4.4 &71.7 {\tiny(3.8)} / 73.0 {\tiny(5.4)}& 69.5 {\tiny (4.2)} \\\midrule
 {\sysname} w/o Adapter ($355M$) & 72.6 & 2.5 & 73.6 {\tiny(2.7)} / 74.8 {\tiny(2.7)} & 71.2 {\tiny(2.3)}\\
\bottomrule
\end{tabular}
}
\caption{Ablation analysis of {\sysname} with 30 labels on MNLI and RTE with tunable parameters in parantheses.} \vspace{-0.15in}
\label{tab:ablation_model}
\end{table}

\vspace{-0.0in}

\noindent$\bullet$ \textbf{Knowledge Distillation Warmup.} In this ablation study, we remove the warmup phase with knowledge distillation from {\tt LiST} (denoted as ``LiST w/o KD Warmup''). Removing this component results in $4\%$ performance drop in terms of average accuracy and $300\%$ larger standard deviation -- demonstrating the importance of KD Warmup in stabilizing {\tt LiST} training.

\vspace{-0.0in}

\noindent$\bullet$ \textbf{{\sysname} versus {\sysname} w/o Adapter.} In {\tt LiST}, we only fine-tune the adapter and language model head while keeping other encoder parameters frozen to achieve parameter efficiency. 
Table~\ref{tab:ablation_model} shows that {\sysname} using only $4\%$ tunable parameters is able to match the performance of fully tunable {\sysname} (that is without using any adapters and tuning all encoder parameters) on MNLI and RTE -- demonstrating the effectiveness of our lightweight design.  {More ablation results with varying shots are presented in Appendix in Tables~\ref{tab_app:ablation_30},~\ref{tab_app:ablation_20} and~\ref{tab_app:ablation_10} that demonstrate similar trends as in Table~\ref{tab:ablation_model}.}



\vspace{-0.05in}
\section{Conclusions and Future Work}
\vspace{-0.05in}
We develop a new method {\tt LiST} for lightweight tuning of large language models in few-shot settings. {\sysname} uses prompted self-training to learn from large amounts of unlabeled data from target domains. In order to reduce the storage and training cost, {\sysname} tunes only a small number of adapter parameters with few-shot labels while keeping the large encoder frozen. With only $30$ labels for every task, {\sysname} improves by upto $35\%$ over classic fine-tuning and $6\%$ over prompt-tuning while reducing $96\%$ of the tunable parameters. With significant reduction in the cost of (data) annotation and overall model footprint, {\sysname} provides an efficient framework towards life-long learning of AI agents~\citep{biesialska2020continual}. 
While adapters reduce storage cost, {\sysname} does not reduce inference latency given the PLM backbone. A future work is to consider combining model compression techniques~\citep{han2015learning,cai2019once} with adapters to reduce FLOPS and latency.




\section{Ethical Considerations}
In this work, we introduce a lightweight framework for self-training of language models with only a few labeled examples. We expect that progress and findings presented in this paper could further benefit NLP applications with limited labeled data. In the real-world setting, it is usually not only expensive to obtain large-scale labeled data for each task but also brings privacy and compliance concerns when large-scale data labeling is needed. The privacy concerns could be further exacerbated when dealing with sensitive user data for peronslization tasks. Our framework which only needs few-shot labeled data could help in this  to obtain state-of-the-art-performance while alleviating privacy concerns. The proposed framework is tested across different tasks and could be used  for applications in various areas including finance, legal, healthcare, retail and other domains where adoption of deep neural network may have been hindered due to lack of large-scale manual annotations on sensitive user data.

While our framework advance the progress of NLP, it also suffers from associated societal implications of automation ranging from job losses for workers who provide annotations as a service as well as for other industries relying on human labor. Moreover, it may bring additional concerns when  NLP models are used by malicious agents for propagating bias, misinformation and indulging in other nefarious activities. However, many of these concerns can also be alleviated with our framework to develop better detection models and mitigation strategies with only a few representative examples of such intents.

The proposed method is somewhat compute-intensive as it involves large-scale language model. This might impose negative impact on carbon footprint from training the described models.  In order to reduce the storage and training cost, the proposed design  tunes only a small number of adapter parameters with few-shot labels while keeping the large encoder frozen.

\bibliography{anthology,custom}

\begin{thebibliography}{49}
\expandafter\ifx\csname natexlab\endcsname\relax\def\natexlab#1{#1}\fi

\bibitem[{Aghajanyan et~al.(2021)Aghajanyan, Gupta, and
  Zettlemoyer}]{aghajanyan2020intrinsic}
Armen Aghajanyan, Sonal Gupta, and Luke Zettlemoyer. 2021.
\newblock \href {https://doi.org/10.18653/v1/2021.acl-long.568} {Intrinsic
  dimensionality explains the effectiveness of language model fine-tuning}.
\newblock In \emph{Proceedings of the 59th Annual Meeting of the Association
  for Computational Linguistics and the 11th International Joint Conference on
  Natural Language Processing (Volume 1: Long Papers)}, pages 7319--7328,
  Online. Association for Computational Linguistics.

\bibitem[{Bar~Haim et~al.(2006)Bar~Haim, Dagan, Dolan, Ferro, Giampiccolo,
  Magnini, and Szpektor}]{bar2006second}
Roy Bar~Haim, Ido Dagan, Bill Dolan, Lisa Ferro, Danilo Giampiccolo, Bernardo
  Magnini, and Idan Szpektor. 2006.
\newblock The second {PASCAL} recognising textual entailment challenge.

\bibitem[{Beck et~al.(2021)Beck, Bohlender, Viehmann, Hane, Adamson, Khuri,
  Brossmann, Pfeiffer, and Gurevych}]{beck2021adapterhub}
Tilman Beck, Bela Bohlender, Christina Viehmann, Vincent Hane, Yanik Adamson,
  Jaber Khuri, Jonas Brossmann, Jonas Pfeiffer, and Iryna Gurevych. 2021.
\newblock Adapterhub playground: Simple and flexible few-shot learning with
  adapters.
\newblock \emph{arXiv preprint arXiv:2108.08103}.

\bibitem[{Bentivogli et~al.(2009)Bentivogli, Clark, Dagan, and
  Giampiccolo}]{bentivogli2009fifth-rte4}
Luisa Bentivogli, Peter Clark, Ido Dagan, and Danilo Giampiccolo. 2009.
\newblock The fifth {PASCAL} recognizing textual entailment challenge.
\newblock In \emph{TAC}.

\bibitem[{Biesialska et~al.(2020)Biesialska, Biesialska, and
  Costa-juss{\`a}}]{biesialska2020continual}
Magdalena Biesialska, Katarzyna Biesialska, and Marta~R Costa-juss{\`a}. 2020.
\newblock Continual lifelong learning in natural language processing: A survey.
\newblock In \emph{Proceedings of the 28th International Conference on
  Computational Linguistics}, pages 6523--6541.

\bibitem[{Brown et~al.(2020)Brown, Mann, Ryder, Subbiah, Kaplan, Dhariwal,
  Neelakantan, Shyam, Sastry, Askell, Agarwal, Herbert-Voss, Krueger, Henighan,
  Child, Ramesh, Ziegler, Wu, Winter, Hesse, Chen, Sigler, Litwin, Gray, Chess,
  Clark, Berner, McCandlish, Radford, Sutskever, and
  Amodei}]{brown2020language}
Tom Brown, Benjamin Mann, Nick Ryder, Melanie Subbiah, Jared~D Kaplan, Prafulla
  Dhariwal, Arvind Neelakantan, Pranav Shyam, Girish Sastry, Amanda Askell,
  Sandhini Agarwal, Ariel Herbert-Voss, Gretchen Krueger, Tom Henighan, Rewon
  Child, Aditya Ramesh, Daniel Ziegler, Jeffrey Wu, Clemens Winter, Chris
  Hesse, Mark Chen, Eric Sigler, Mateusz Litwin, Scott Gray, Benjamin Chess,
  Jack Clark, Christopher Berner, Sam McCandlish, Alec Radford, Ilya Sutskever,
  and Dario Amodei. 2020.
\newblock \href
  {https://proceedings.neurips.cc/paper/2020/file/1457c0d6bfcb4967418bfb8ac142f64a-Paper.pdf}
  {Language models are few-shot learners}.
\newblock In \emph{Advances in Neural Information Processing Systems},
  volume~33, pages 1877--1901. Curran Associates, Inc.

\bibitem[{Cai et~al.(2020{\natexlab{a}})Cai, Gan, Wang, Zhang, and
  Han}]{cai2019once}
Han Cai, Chuang Gan, Tianzhe Wang, Zhekai Zhang, and Song Han.
  2020{\natexlab{a}}.
\newblock Once-for-all: Train one network and specialize it for efficient
  deployment.
\newblock In \emph{International Conference on Learning Representations}.

\bibitem[{Cai et~al.(2020{\natexlab{b}})Cai, Gan, Zhu, and Han}]{cai2020tinytl}
Han Cai, Chuang Gan, Ligeng Zhu, and Song Han. 2020{\natexlab{b}}.
\newblock Tinytl: Reduce memory, not parameters for efficient on-device
  learning.
\newblock \emph{Advances in Neural Information Processing Systems}, 33.

\bibitem[{Clark et~al.(2020)Clark, Luong, Le, and
  Manning}]{DBLP:conf/iclr/ClarkLLM20}
Kevin Clark, Minh{-}Thang Luong, Quoc~V. Le, and Christopher~D. Manning. 2020.
\newblock \href {https://openreview.net/forum?id=r1xMH1BtvB} {{ELECTRA:}
  pre-training text encoders as discriminators rather than generators}.
\newblock In \emph{8th International Conference on Learning Representations,
  {ICLR} 2020, Addis Ababa, Ethiopia, April 26-30, 2020}. OpenReview.net.

\bibitem[{Dagan et~al.(2005)Dagan, Glickman, and
  Magnini}]{dagan2005pascal-rte1}
Ido Dagan, Oren Glickman, and Bernardo Magnini. 2005.
\newblock The {PASCAL} recognising textual entailment challenge.
\newblock In \emph{the First International Conference on Machine Learning
  Challenges: Evaluating Predictive Uncertainty Visual Object Classification,
  and Recognizing Textual Entailment}.

\bibitem[{Devlin et~al.(2019{\natexlab{a}})Devlin, Chang, Lee, and
  Toutanova}]{DBLP:conf/naacl/DevlinCLT19}
Jacob Devlin, Ming{-}Wei Chang, Kenton Lee, and Kristina Toutanova.
  2019{\natexlab{a}}.
\newblock \href {https://aclweb.org/anthology/papers/N/N19/N19-1423/} {{BERT:}
  pre-training of deep bidirectional transformers for language understanding}.
\newblock In \emph{Proceedings of the 2019 Conference of the North American
  Chapter of the Association for Computational Linguistics: Human Language
  Technologies, {NAACL-HLT} 2019, Minneapolis, MN, USA, June 2-7, 2019, Volume
  1 (Long and Short Papers)}, pages 4171--4186.

\bibitem[{Devlin et~al.(2019{\natexlab{b}})Devlin, Chang, Lee, and
  Toutanova}]{BERT}
Jacob Devlin, Ming{-}Wei Chang, Kenton Lee, and Kristina Toutanova.
  2019{\natexlab{b}}.
\newblock {BERT:} pre-training of deep bidirectional transformers for language
  understanding.
\newblock In \emph{Proceedings of the 2019 Conference of the North American
  Chapter of the Association for Computational Linguistics: Human Language
  Technologies, {NAACL-HLT} 2019, Volume 1 (Long and Short Papers)}, pages
  4171--4186.

\bibitem[{Finn et~al.(2017)Finn, Abbeel, and Levine}]{finn2017model}
Chelsea Finn, Pieter Abbeel, and Sergey Levine. 2017.
\newblock Model-agnostic meta-learning for fast adaptation of deep networks.
\newblock In \emph{International Conference on Machine Learning}, pages
  1126--1135. PMLR.

\bibitem[{Gao et~al.(2021)Gao, Fisch, and Chen}]{gao2021making}
Tianyu Gao, Adam Fisch, and Danqi Chen. 2021.
\newblock Making pre-trained language models better few-shot learners.
\newblock In \emph{Association for Computational Linguistics (ACL)}.

\bibitem[{Giampiccolo et~al.(2007)Giampiccolo, Magnini, Dagan, and
  Dolan}]{giampiccolo2007third-rte3}
Danilo Giampiccolo, Bernardo Magnini, Ido Dagan, and Bill Dolan. 2007.
\newblock The third {PASCAL} recognizing textual entailment challenge.
\newblock In \emph{the {ACL}-{PASCAL} Workshop on Textual Entailment and
  Paraphrasing}.

\bibitem[{Gunel et~al.(2020)Gunel, Du, Conneau, and
  Stoyanov}]{gunel2020supervised}
Beliz Gunel, Jingfei Du, Alexis Conneau, and Veselin Stoyanov. 2020.
\newblock Supervised contrastive learning for pre-trained language model
  fine-tuning.
\newblock In \emph{International Conference on Learning Representations}.

\bibitem[{Han et~al.(2015)Han, Pool, Tran, and Dally}]{han2015learning}
Song Han, Jeff Pool, John Tran, and William~J Dally. 2015.
\newblock Learning both weights and connections for efficient neural network.
\newblock In \emph{NIPS}.

\bibitem[{He et~al.(2019)He, Gu, Shen, and Ranzato}]{he2019revisiting}
Junxian He, Jiatao Gu, Jiajun Shen, and Marc'Aurelio Ranzato. 2019.
\newblock \href {http://arxiv.org/abs/1909.13788} {Revisiting self-training for
  neural sequence generation}.

\bibitem[{He et~al.(2021)He, Liu, Gao, and Chen}]{DBLP:conf/iclr/HeLGC21}
Pengcheng He, Xiaodong Liu, Jianfeng Gao, and Weizhu Chen. 2021.
\newblock \href {https://openreview.net/forum?id=XPZIaotutsD} {Deberta:
  decoding-enhanced bert with disentangled attention}.
\newblock In \emph{9th International Conference on Learning Representations,
  {ICLR} 2021, Virtual Event, Austria, May 3-7, 2021}. OpenReview.net.

\bibitem[{Hinton et~al.(2015)Hinton, Vinyals, and Dean}]{hinton2015distilling}
Geoffrey Hinton, Oriol Vinyals, and Jeff Dean. 2015.
\newblock Distilling the knowledge in a neural network.
\newblock \emph{arXiv preprint arXiv:1503.02531}.

\bibitem[{Houlsby et~al.(2019)Houlsby, Giurgiu, Jastrzebski, Morrone,
  De~Laroussilhe, Gesmundo, Attariyan, and Gelly}]{houlsby2019parameter}
Neil Houlsby, Andrei Giurgiu, Stanislaw Jastrzebski, Bruna Morrone, Quentin
  De~Laroussilhe, Andrea Gesmundo, Mona Attariyan, and Sylvain Gelly. 2019.
\newblock Parameter-efficient transfer learning for nlp.
\newblock In \emph{International Conference on Machine Learning}, pages
  2790--2799. PMLR.

\bibitem[{LeCun et~al.(1990)LeCun, Denker, and Solla}]{lecun1990optimal}
Yann LeCun, John~S Denker, and Sara~A Solla. 1990.
\newblock Optimal brain damage.
\newblock In \emph{Advances in neural information processing systems}, pages
  598--605.

\bibitem[{Lester et~al.(2021{\natexlab{a}})Lester, Al{-}Rfou, and
  Constant}]{prompt_tuning}
Brian Lester, Rami Al{-}Rfou, and Noah Constant. 2021{\natexlab{a}}.
\newblock \href {http://arxiv.org/abs/2104.08691} {The power of scale for
  parameter-efficient prompt tuning}.
\newblock \emph{CoRR}, abs/2104.08691.

\bibitem[{Lester et~al.(2021{\natexlab{b}})Lester, Al-Rfou, and
  Constant}]{lester2021power}
Brian Lester, Rami Al-Rfou, and Noah Constant. 2021{\natexlab{b}}.
\newblock The power of scale for parameter-efficient prompt tuning.
\newblock \emph{arXiv preprint arXiv:2104.08691}.

\bibitem[{Li and Liang(2021)}]{prefix}
Xiang~Lisa Li and Percy Liang. 2021.
\newblock \href {http://arxiv.org/abs/2101.00190} {Prefix-tuning: Optimizing
  continuous prompts for generation}.
\newblock \emph{CoRR}, abs/2101.00190.

\bibitem[{Li et~al.(2019)Li, Sun, Liu, Zhou, Zheng, Chua, and
  Schiele}]{li2019learning}
Xinzhe Li, Qianru Sun, Yaoyao Liu, Qin Zhou, Shibao Zheng, Tat-Seng Chua, and
  Bernt Schiele. 2019.
\newblock Learning to self-train for semi-supervised few-shot classification.
\newblock \emph{Advances in Neural Information Processing Systems},
  32:10276--10286.

\bibitem[{Liu et~al.(2019{\natexlab{a}})Liu, Ott, Goyal, Du, Joshi, Chen, Levy,
  Lewis, Zettlemoyer, and Stoyanov}]{DBLP:journals/corr/abs-1907-11692}
Yinhan Liu, Myle Ott, Naman Goyal, Jingfei Du, Mandar Joshi, Danqi Chen, Omer
  Levy, Mike Lewis, Luke Zettlemoyer, and Veselin Stoyanov. 2019{\natexlab{a}}.
\newblock \href {http://arxiv.org/abs/1907.11692} {Roberta: {A} robustly
  optimized {BERT} pretraining approach}.
\newblock \emph{CoRR}, abs/1907.11692.

\bibitem[{Liu et~al.(2019{\natexlab{b}})Liu, Ott, Goyal, Du, Joshi, Chen, Levy,
  Lewis, Zettlemoyer, and Stoyanov}]{RoBERTa}
Yinhan Liu, Myle Ott, Naman Goyal, Jingfei Du, Mandar Joshi, Danqi Chen, Omer
  Levy, Mike Lewis, Luke Zettlemoyer, and Veselin Stoyanov. 2019{\natexlab{b}}.
\newblock \href {http://arxiv.org/abs/1907.11692} {Roberta: {A} robustly
  optimized {BERT} pretraining approach}.
\newblock \emph{CoRR}, abs/1907.11692.

\bibitem[{Loshchilov and Hutter(2017)}]{loshchilov2017decoupled}
Ilya Loshchilov and Frank Hutter. 2017.
\newblock Decoupled weight decay regularization.
\newblock \emph{arXiv preprint arXiv:1711.05101}.

\bibitem[{Mukherjee and Awadallah(2020)}]{mukherjee2020uncertainty}
Subhabrata Mukherjee and Ahmed Awadallah. 2020.
\newblock Uncertainty-aware self-training for few-shot text classification.
\newblock \emph{Advances in Neural Information Processing Systems}, 33.

\bibitem[{Pang and Lee(2004)}]{pang2004sentimental-subj}
Bo~Pang and Lillian Lee. 2004.
\newblock A sentimental education: Sentiment analysis using subjectivity
  summarization based on minimum cuts.

\bibitem[{Perez et~al.(2021)Perez, Kiela, and Cho}]{perez2021true}
Ethan Perez, Douwe Kiela, and Kyunghyun Cho. 2021.
\newblock True few-shot learning with language models.
\newblock \emph{arXiv preprint arXiv:2105.11447}.

\bibitem[{Pfeiffer et~al.(2020)Pfeiffer, R\"uckl\'{e}, Poth, Kamath, Vuli\'{c},
  Ruder, Cho, and Gurevych}]{pfeiffer2020AdapterHub}
Jonas Pfeiffer, Andreas R\"uckl\'{e}, Clifton Poth, Aishwarya Kamath, Ivan
  Vuli\'{c}, Sebastian Ruder, Kyunghyun Cho, and Iryna Gurevych. 2020.
\newblock \href {https://www.aclweb.org/anthology/2020.emnlp-demos.7}
  {Adapterhub: A framework for adapting transformers}.
\newblock In \emph{Proceedings of the 2020 Conference on Empirical Methods in
  Natural Language Processing (EMNLP 2020): Systems Demonstrations}, pages
  46--54, Online. Association for Computational Linguistics.

\bibitem[{Raffel et~al.(2020)Raffel, Shazeer, Roberts, Lee, Narang, Matena,
  Zhou, Li, and Liu}]{T5}
Colin Raffel, Noam Shazeer, Adam Roberts, Katherine Lee, Sharan Narang, Michael
  Matena, Yanqi Zhou, Wei Li, and Peter~J Liu. 2020.
\newblock Exploring the limits of transfer learning with a unified text-to-text
  transformer.
\newblock \emph{Journal of Machine Learning Research}, 21(140):1--67.

\bibitem[{Ren et~al.(2018)Ren, Zeng, Yang, and Urtasun}]{ren2018learning}
Mengye Ren, Wenyuan Zeng, Bin Yang, and Raquel Urtasun. 2018.
\newblock Learning to reweight examples for robust deep learning.
\newblock In \emph{International Conference on Machine Learning}, pages
  4334--4343. PMLR.

\bibitem[{Schick and Sch{\"u}tze(2021{\natexlab{a}})}]{schick2021exploiting}
Timo Schick and Hinrich Sch{\"u}tze. 2021{\natexlab{a}}.
\newblock Exploiting cloze-questions for few-shot text classification and
  natural language inference.
\newblock In \emph{Proceedings of the 16th Conference of the European Chapter
  of the Association for Computational Linguistics: Main Volume}, pages
  255--269.

\bibitem[{Schick and
  Sch{\"u}tze(2021{\natexlab{b}})}]{schick-schutze-2021-just}
Timo Schick and Hinrich Sch{\"u}tze. 2021{\natexlab{b}}.
\newblock \href {https://doi.org/10.18653/v1/2021.naacl-main.185} {It{'}s not
  just size that matters: Small language models are also few-shot learners}.
\newblock In \emph{Proceedings of the 2021 Conference of the North American
  Chapter of the Association for Computational Linguistics: Human Language
  Technologies}, pages 2339--2352, Online. Association for Computational
  Linguistics.

\bibitem[{Socher et~al.()Socher, Perelygin, Wu, Chuang, Manning, Ng, and
  Potts}]{socher2013recursive-sst-2}
Richard Socher, Alex Perelygin, Jean Wu, Jason Chuang, Christopher~D. Manning,
  Andrew Ng, and Christopher Potts.
\newblock Recursive deep models for semantic compositionality over a sentiment
  treebank.

\bibitem[{Vu et~al.(2021)Vu, Luong, Le, Simon, and Iyyer}]{vu2021strata}
Tu~Vu, Minh-Thang Luong, Quoc~V Le, Grady Simon, and Mohit Iyyer. 2021.
\newblock Strata: Self-training with task augmentation for better few-shot
  learning.
\newblock \emph{arXiv preprint arXiv:2109.06270}.

\bibitem[{Wang et~al.(2019)Wang, Singh, Michael, Hill, Levy, and
  Bowman}]{wang2019glue}
Alex Wang, Amanpreet Singh, Julian Michael, Felix Hill, Omer Levy, and Samuel~R
  Bowman. 2019.
\newblock {GLUE}: A multi-task benchmark and analysis platform for natural
  language understanding.

\bibitem[{Wang et~al.(2021{\natexlab{a}})Wang, Chu, Zhang, and
  Gao}]{wang2021learning}
Yaqing Wang, Haoda Chu, Chao Zhang, and Jing Gao. 2021{\natexlab{a}}.
\newblock Learning from language description: Low-shot named entity recognition
  via decomposed framework.
\newblock In \emph{Findings of the Association for Computational Linguistics:
  EMNLP 2021}, pages 1618--1630.

\bibitem[{Wang et~al.(2021{\natexlab{b}})Wang, Ma, Wang, Jha, and
  Gao}]{wang2021multimodal}
Yaqing Wang, Fenglong Ma, Haoyu Wang, Kishlay Jha, and Jing Gao.
  2021{\natexlab{b}}.
\newblock Multimodal emergent fake news detection via meta neural process
  networks.
\newblock In \emph{Proceedings of the 27th ACM SIGKDD Conference on Knowledge
  Discovery \& Data Mining}, pages 3708--3716.

\bibitem[{Wang et~al.(2021{\natexlab{c}})Wang, Mukherjee, Chu, Tu, Wu, Gao, and
  Awadallah}]{wang2021meta}
Yaqing Wang, Subhabrata Mukherjee, Haoda Chu, Yuancheng Tu, Ming Wu, Jing Gao,
  and Ahmed~Hassan Awadallah. 2021{\natexlab{c}}.
\newblock Meta self-training for few-shot neural sequence labeling.
\newblock In \emph{Proceedings of the 27th ACM SIGKDD Conference on Knowledge
  Discovery \& Data Mining}, pages 1737--1747.

\bibitem[{Wiebe et~al.(2005)Wiebe, Wilson, and Cardie}]{wiebe2005annotating}
Janyce Wiebe, Theresa Wilson, and Claire Cardie. 2005.
\newblock Annotating expressions of opinions and emotions in language.
\newblock \emph{Language resources and evaluation}, 39(2):165--210.

\bibitem[{Williams et~al.(2018{\natexlab{a}})Williams, Nangia, and
  Bowman}]{mnli}
Adina Williams, Nikita Nangia, and Samuel Bowman. 2018{\natexlab{a}}.
\newblock \href {http://aclweb.org/anthology/N18-1101} {A broad-coverage
  challenge corpus for sentence understanding through inference}.
\newblock In \emph{Proceedings of the 2018 Conference of the North American
  Chapter of the Association for Computational Linguistics: Human Language
  Technologies, Volume 1 (Long Papers)}, pages 1112--1122. Association for
  Computational Linguistics.

\bibitem[{Williams et~al.(2018{\natexlab{b}})Williams, Nangia, and
  Bowman}]{williams2018broad-mnli}
Adina Williams, Nikita Nangia, and Samuel Bowman. 2018{\natexlab{b}}.
\newblock A broad-coverage challenge corpus for sentence understanding through
  inference.

\bibitem[{Xie et~al.(2019)Xie, Dai, Hovy, Luong, and Le}]{xie2019unsupervised}
Qizhe Xie, Zihang Dai, Eduard Hovy, Minh-Thang Luong, and Quoc~V Le. 2019.
\newblock Unsupervised data augmentation for consistency training.
\newblock \emph{arXiv preprint arXiv:1904.12848}.

\bibitem[{Zhang et~al.(2021)Zhang, Wu, Katiyar, Weinberger, and
  Artzi}]{zhang2020revisiting}
Tianyi Zhang, Felix Wu, Arzoo Katiyar, Kilian~Q Weinberger, and Yoav Artzi.
  2021.
\newblock Revisiting few-sample {BERT} fine-tuning.

\bibitem[{Zhong et~al.(2021)Zhong, Tang, Wang, Yin, and
  Duan}]{zhong2021useradapter}
Wanjun Zhong, Duyu Tang, Jiahai Wang, Jian Yin, and Nan Duan. 2021.
\newblock Useradapter: Few-shot user learning in sentiment analysis.
\newblock In \emph{ACL/IJCNLP (Findings)}.

\end{thebibliography}
\bibliographystyle{acl_natbib}
\clearpage
\appendix

\appendix
\section{Datasets}
\label{sec:data}

\subsection{Dataset information}

Table~\ref{tab:datasets} summarize dataset statistics and task descriptions. All the datasets are in English Language. The licence information is as follows.

MNLI: The majority of the corpus is released under the OANC’s license, which allows all content to be freely used, modified, and shared under permissive terms. The data in the FICTION section falls under several permissive licenses; Seven Swords is available under a Creative Commons Share-Alike 3.0 Unported License, and with the explicit permission of the author, Living History and Password Incorrect are available under Creative Commons Attribution 3.0 Unported Licenses; the remaining works of fiction are in the public domain in the United States (but may be licensed differently elsewhere).

RTE: The dataset is public release but the corresponding licence information is not found in the source website~\footnote{\url{https://aclweb.org/aclwiki/Recognizing_Textual_Entailment}}. 

QQP: We did not find the responding license. The source website~\footnote{\url{https://data.quora.com/First-Quora-Dataset-Release-Question-Pairs}} is not accessible. 

SST-2 dataset:  CC0: Public Domain

Subj: The dataset is public release but the Licence information is not presented in the source website~\footnote{\url{http://www.cs.cornell.edu/people/pabo/movie-review-data/}}.

MPQA: The dataset\footnote{\url{https://mpqa.cs.pitt.edu/}} is public release. Made available under the terms of GNU General Public License. They are distributed without any warranty.

We follow the licence of datasets for research use.  We manually check no offensive content in our few-shot training dataset. 

\begin{table*}[h]
\begin{center}
\small
\vspace{-0.1in}
\centering
\resizebox{0.8\textwidth}{!}{%
\begin{tabular}{llcrccl}
\toprule
\textbf{Category} & \tf{Dataset} & \#Labels &  \tf{\#Full Train} & \tf{\#Test} & \tf{Type} & \tf{Labels} \\
\midrule
\multirow{3}{*}{\thead{sentence- \\pair  }} & MNLI & 3 &  392,702 & 9,815 & NLI & entailment, neutral, contradiction\\
& RTE & 2 &   2,490 & 277 & NLI &  entailment, not\_entailment \\
& QQP & 2 & 363,846 & 40,431 & paraphrase & equivalent, not\_equivalent  \\
\midrule
  \multirow{3}{*}{\thead{single-\\sentence }} & SST-2 & 2 &  6,920 & 872 & sentiment & positive, negative \\
 & Subj & 2 &  8,000 & 2,000 & subjectivity & subjective, objective \\
  & MPQA & 2 & 8,606 & 2,000 & opinion polarity & positive, negative \\
\bottomrule
\end{tabular}
}
\end{center}
\caption{Dataset summary and task descriptions. For each task, we sample $\mathcal{K} \in \{10, 20, 30\}$ labeled examples to form five different splits with different random seeds from the original training set, and add the remaining to the unlabeled set while ignoring their labels.}
\label{tab:datasets}
\end{table*}



\subsection{Prompts}
Table ~\ref{tab:manual_prompts} summarizes manually-designed prompts and label words  for each
dataset in our experiments. These prompts and label words were adopted from~\citep{gao2021making}.

\begin{table*}[h]
\begin{center}
\centering
\resizebox{0.8\linewidth}{!}{%
\begin{tabular}{lll}
\toprule
\tf{Task} & \tf{Prompt} & \tf{Label words}\\
\midrule
SST-2 &  {\sent} It was {\mask} . & positive: great, negative: terrible\\
MR    & {\sent} It was {\mask} . & positive: great, negative: terrible\\

Subj  & {\sent} This is {\mask} . & subjective: subjective, objective: objective \\
\midrule
MNLI  & {\firstsent} ? {\mask} , {\secondsent} & entailment: Yes, netural: Maybe, contradiction: No \\
RTE   & {\firstsent} ? {\mask} , {\secondsent} & entailment: Yes, not\_entailment: No \\
QQP   & {\firstsent} {\mask} , {\secondsent} & equivalent: Yes, not\_equivalent: No\\
\bottomrule
\end{tabular}
}
\end{center}
\caption{Task prompt and label words summary. {\sent} and {\secondsent} indicate input sentences.
}
\label{tab:manual_prompts}
\end{table*}

\section{Algorithm Flow}
\label{sec:alg}

Algorithm~\ref{alg} summarizes overall flow of {\tt LiST}. We adopt a light self-training mechanism which keeps the shared student and teacher encoder parameters frozen and only updates the adapter parameters along with the corresponding language model heads. Beside the lightweight tuning design, another key step in our self-training framework is to utilize the few-shot labeled data to fine-tune the 
student model $\psi^{(T)}_{\mathrm{stu}}$) in every self-training session. Such a step is different with conventional self-training framework, which either leverages labeled data for initial teacher fine-tuning or combine labeled data with unlabeled data for joint training of student model.  The iterative usage of unlabeled data and labeled data helps in better teacher initialization before next round of adapter prompt-tuning on $\widetilde{\mathcal{D}}^{Train}_K$ which further helps in improving model tuning and the quality of pseudo labels.

\begin{algorithm}[h]
\caption{{\sysname} Algorithm.}
\SetAlgoLined
\scriptsize
\KwIn{Labeled samples $\widetilde{\mathcal{D}}^{Train}_K=\{\widetilde{x}^l, \widetilde{y}^l\}$; Unlabeled samples $\widetilde{\mathcal{D}}^U=\{\widetilde{x}_u\}$; a  pre-trained language model with parameters $\Theta_{\mathrm{PLM}}$; randomly initialized Adapter with parameters $\psi$;  Number of student training iterations $T$, KD warmup steps $T_{warm}$ and self-training sessions $M$.}
\parbox[t]{.7\textwidth}{Initialize teacher adapter  $\psi_{\mathrm{tea}} = \psi^{(0)}$ }

Tune teacher adapter $\psi_{\mathrm{tea}}$  on small labeled data $\widetilde{\mathcal{D}}^{Train}_K$;

  \For{$m \gets1$ \KwTo $M$}{

  Initialize the student adapter $\psi_{\mathrm{stu}} = \psi^{(0)}$ ;
  
   \For{$t\gets1$ \KwTo $T$}{
      Infer pseudo prompt labels $\{\hat{y}^{(t)}_{n}\}_{n=1}^N$ for unlabeled data $\widetilde{\mathcal{D}}^U=\{\widetilde{x}_u\}$ with teacher model $(\Theta_{\mathrm{PLM}}, \psi_{\mathrm{tea}})$;
 
      Randomly sample a batch of pseudo-labeled samples from $(\widetilde{x}_u,\hat{y}^{(t)})$;
      
   \eIf{t $<$ T$_{warm}$}{ 
   Train student adapter $\psi_{\mathrm{stu}}$ according to Eq.~\ref{Eq:kd_warmup}
   }{
   
   \parbox[t]{.6\linewidth}{Sample a mini-batch  from $\widetilde{\mathcal{D}}^{(t)} \in \widetilde{\mathcal{D}}^{Train}_K$ as validation mini-batc for re-weighting;}
   
     Train student adapter $\psi_{\mathrm{stu}}$ on re-weighted  pseudo-labeled samples according to Eq.~\ref{Eq:meta_update};
    
   }
   
    }
    
    Tune student adapter $\psi^{(T)}_{\mathrm{stu}}$  on small labeled data $\widetilde{\mathcal{D}}^{Train}_K$;
    
 Update the teacher adapter: $\psi_{\mathrm{tea}} = \psi_{\mathrm{stu}}^{(T)}$
   
 }\label{alg}
\end{algorithm}

\section{Experimental Details}

\subsection{Hyper-parameters}
\label{subsec:hyper}

Following the true few-shot learning spirit, we do not have any additional development set for hyper-parameter tuning. Instead we keep all the hyper-parameter same for different tasks, different model families and sizes as well as different shots $K$. We retain most of the default hyper-parameter configurations from related work. For each task, we run the model five times with different data splits and different random seeds in $\{1,2,3,4,5\}$. Our experiments are conducted in few-shot supervision setting and few-shot semi-supervised setting. In the following, we introduce the hyper-parameters for each setting respectively.   

\noindent\textbf{Few-shot supervision setting.} We set learning rate as 5e-6, training epochs as $400$ and batch size as $4$. The bottleneck dimension $d$ of Adapter is set to $128$. The optimizer is AdamW~\citep{loshchilov2017decoupled} with default settings besides learning rate. {We use variance for adapter as 0.002 and observe that the performance is not sensitive to variance values when the scale of variance values are equal or less than 0.002. Since experiments are run with different number of labeled examples,  the GPU hours range from 5 minutes to 1 hour per task. }

\noindent\textbf{Few-shot semi-supervised setting.} For initial teacher fine-tuning, we adopt the same hyper-parameter configuration as in few-shot supervision setting. To facilitate training on a large amounts of unlabeled data, the learning rate in self-training is set to 1e-4 following fully supervised adapter work~\citep{pfeiffer2020AdapterHub}. The batch size of unlabeled data for student adapter training is $16$ and the size of minibatch $\widetilde{\mathcal{D}} \in \widetilde{\mathcal{D}}^{Train}_K$ for meta re-weighiting in Eq.~\ref{eq:grad} is $4$. For each self-training session, we train student adapter for $T=1000$ steps and further fine-tune $50$ epochs on given labeled data. The student KD warmup ratio is set to $60\%$, i.e., $T_{warm}=600$ steps, without extra hyper-parameter tuning. We repeat all the steps in self-training training $M=6$ times.  Since experiments are run with different number of labeled examples and datasets,  the GPU hours of all approaches are different, ranging from 1 hour to 10 hours per task.

\begin{table}[h]
\centering
\resizebox{0.7\linewidth}{!}{
\begin{tabular}{clc}
\toprule
 \textbf{Models}  & \textbf{\#Params} &\textbf{Avg Acc} \\
\midrule

BERT-base & 110M & 67.4  \\

BERT-large  &  336M  & 68.0\\
\midrule
 RoBERTa-base  &125M & 73.7  \\

 RoBERTa-large  & 355M & \textbf{77.6} \\

\midrule

T5-small   & 60M & 66.5 \\

T5-base & 220M  & 71.9 
\\

T5-large  &770M & 77.3 \\

\midrule

\end{tabular}
}\vspace{-0.5em}
\caption{Average accuracy of prompt FN with different encoders using $|\mathcal{K}|=30$ labels  on six tasks.}
\vspace{-0.1in}
\label{tab:var_model}
\end{table}

\vspace{-0.1in}
\subsection{Few-shot Supervision with Varying Model Sizes and Labels}\vspace{-0.5em}

To better understand the role of different model families in few-shot prompt-based FN, we evaluate the performance of representative state-of-the-art PLMs like BERT~\citep{BERT}, RoBERTa~\citep{RoBERTa} and T5~\citep{T5} of different sizes (parameters) using varying amounts of labeled data. 
We report macro-averaged results over six tasks where each has five different splits  for easy comparison.

\noindent\textbf{Effect of model choices.}  Table~\ref{tab:var_model} shows the performance comparison of three representative PLMs with different parameters using prompt-based FN on $30$ labeled samples. We observe that average performance increases with increase in model size within each model family. Overall, we observe RoBERTa models to perform much better than BERT, and marginally outperform T5 models of much bigger size. Accordingly, we use RoBERTa-large as the base encoder for both {\sysname} and other baseline methods.


\noindent\textbf{Effect of varying the number of labels $|\mathcal{K}|$.} 
From Figure ~\ref{fig:main_fig}, we observe prompt-based FN to consistently outperform classic-tuning under all labeled data settings when using the same encoder. With increase in the amount of labeled examples, prompt-based FN and classic-tuning both improve in performance, although with reduced performance gap. This demonstrates prompt-based FN to be the most impactful for low-resource settings with few training labels. {\sysname} improves over both classic and prompt-based FN in all settings with massive reduction in number of tunable parameters.

\subsection{Experimental result details}
\label{subsec:more_experiments}

\noindent\textbf{Fine-tuning strategies with varying number of shots.} Table~\ref{tab_app:var_shot} shows the performance comparison of RoBERTa-large with two fine-tuning strategies and varying number of labeled samples including zero-shot supervision, few-shot supervision from 10 to 30 and full supervision. Prompt fine-tuning shows competitive performance in zero-shot learning, outperforming classic fine-tuning strategy with 30 labeled examples on several tasks like MNLI and SST-2. As the size of labeled examples increases, the average performance of classic and prompt fine-tuning strategy improves significantly and  prompt fine-tuning strategy consistently improves classic fine-tuning with a big gap in the few-shot setting. With full supervision, Prompt fine-tuning strategy and classic fine-tuning strategy achieve similar performance, demonstrating that Prompt fine-tuning is most impactful for low-resource settings with few training labels.
 
\begin{table*}[h]
\small
\centering
\resizebox{\textwidth}{!}{
\begin{tabular}{cllllllll}
\toprule
\textbf{Labels} & \textbf{Models} & \textbf{Avg} & \multicolumn{1}{l}{\textbf{MNLI (m/mm)}}  & \multicolumn{1}{l}{\textbf{RTE}}   & \multicolumn{1}{l}{\textbf{QQP}} & \multicolumn{1}{l}{\textbf{SST-2}} & \multicolumn{1}{l}{\textbf{Subj}} & \multicolumn{1}{l}{\textbf{MPQA}}  \\
&&&(acc) &(acc) &(acc) & (acc)& (acc)&(acc)\\
\midrule

\multirow{2}{*}{$|\mathcal{K}| = 0$}
& Classic & - &-  & - & - & - & - & - \\

&Prompt   &58.4  & 51.7/52.4 & 51.3 & 38.6 & 83.6 & 51.4 & 67.6

\\
\midrule

\multirow{2}{*}{$|\mathcal{K}| = 10$}
& Classic  & 50.0 &  34.9 {\tiny(0.3)} / 35.2 {\tiny(0.7)} &50.3 {\tiny(2.1)} & 61.1 {\tiny(3.5)}  &51.8 {\tiny(2.9)} &71.2 {\tiny(17.5)} &52.4 {\tiny(3.2)}\\
& Prompt   &69.3 &  54.8 {\tiny(3.7)} / 55.6 {\tiny(4.6)} & 60.0 {\tiny(4.4)} & 58.7 {\tiny(4.6)} & 89.5 {\tiny(1.7)} & 84.5 {\tiny(8.6)} & 67.8 {\tiny(6.9)}\\
\midrule

\multirow{2}{*}{$|\mathcal{K}| = 20$}
& Classic & 55.2 &   35.8 {\tiny(1.0)} / 36.8 {\tiny(1.5)} & 51.0 {\tiny(4.8)} & 61.3 {\tiny(9.0)}  & 57.2 {\tiny(7.7)} & 84.8 {\tiny(9.0)} & 55.9 {\tiny(4.1)} \\
& Prompt  &75.4 &  60.3 {\tiny(2.0)} / 61.6 {\tiny(2.7)} & 64.3 {\tiny(2.4)} & 67.8 {\tiny(4.2)}& 90.6 {\tiny(1.8)} &  88.3 {\tiny(2.2)} & 80.6 {\tiny(7.5)} \\
\midrule

\multirow{2}{*}{$|\mathcal{K}| = 30$}

&  Classic &59.7 & 38.0 {\tiny(1.7)} / 39.0 {\tiny(3.1)} & 51.4 {\tiny(3.7)} &64.3 {\tiny(8.1)}  & 65.0 {\tiny(11.5)} & 90.2 {\tiny(2.2)} &56.1 {\tiny(5.3)} \\

& Prompt  & 77.6 &  62.8 {\tiny(2.6)} / 64.1 {\tiny(3.3)} & 66.1 {\tiny(2.2)} &  71.1 {\tiny(1.5)} & 91.5 {\tiny(1.0)}& 91.0 {\tiny (0.5)}&82.7 {\tiny(3.8)} \\

\midrule




\multirow{2}{*}{Full supervision}

& Classic &90.7& 89.6 / 89.5 &83.0& 91.8  &95.2&97.2&88.8\\
& Prompt  & 91.8 & 89.3 / 88.8 & 88.4 & 92.1& 95.9& 97.1 & 89.3

 \\
 \midrule
 
\end{tabular}
}
\caption{Average performance and standard deviation of  RoBERTa-large  with Classic and Prompt-tuning strategies with varying training labels
$|\mathcal{K}|$. }
\label{tab_app:var_shot}
\end{table*}

\noindent\textbf{Task performance of varying number of shots and models.} We show performance changes regarding varying number of shots and varying model choices in Figure~\ref{fig:main_fig} and include more detailed results including average accuracy over 5 runs and corresponding standard deviation on MNLI and RTE  in Table~\ref{tab_app:var_shot_model}.

\noindent\textbf{Task performance of different modules with varying number of shots.} We show the average accuracy on tuning different modules of RoBERTa-large with $|K|=30$ on six tasks in Table~\ref{tab:adapter_pilot}. In Table~\ref{tab_app:adapter_pilot}, we show average accuracy with standard deviation of  RoBERTa-large on each task using varying shots of labeled data. We can observe that  Feedforward-output performs best in average while  Attention module  achieves best performance on some tasks. The conclusion is consistent across different shots of labeled data.  Such observations motivate us to insert Adapter into \textit{Feedforward Output} and \textit{Attention} modules to handle diverse tasks.

\noindent\textbf{Task performance of lightweight model tuning strategies.} We show the average accuracy of serveral lightweight strategies with $|K|=30$  labeled examples on six tasks in Table~\ref{tab:adapter}. In Table~\ref{tab_app:adapter}, we show average accuracy with standard deviation of  lightweight tuning strategies on each task with $|K|=30$ labeled examples. We can observe that {\sysname} Adapter outperforms all the lightweight tuning strategies for all six tasks, demonstrating the effective design in adapter placement and parameter initialization. 

\begin{table*}[h]
\small
\centering
\resizebox{0.7\textwidth}{!}{
\begin{tabular}{cllll}
\toprule
\textbf{Labels} & \textbf{Models}  & \multicolumn{1}{l}{\textbf{MNLI (m/mm)}}  & \multicolumn{1}{l}{\textbf{RTE}}    \\
&&(acc) &(acc) \\
\midrule

\multirow{2}{*}{$|\mathcal{K}| = 10$}
& BERT-base-Classic &  32.1 {\tiny(1.2)} / 32.4 {\tiny(1.2)} & 49.3 {\tiny(2.6)}
  \\
& RoBERTa-base-Classic & 35.2 {\tiny(1.1)} / 35.3 {\tiny(1.1)} & 50.6 {\tiny(3.3)}\\
& RoBERTa-large-Classic  & 34.9 {\tiny(0.3)} / 35.2 {\tiny(0.7)} & 50.3 {\tiny(2.1)}
 \\
& BERT-base-Prompt  & 43.0 {\tiny(2.1)} / 44.2 {\tiny(2.1)} & 50.6 {\tiny(3.2)}
 \\
 & RoBERTa-base-Prompt & 49.5 {\tiny(2.9)} / 50.5 {\tiny(3.1)} & 56.5 {\tiny(2.3)} \\

& RoBERTa-large-Prompt & 54.8 {\tiny(3.7)} / 55.6 {\tiny(4.6)} & 59.1 {\tiny(3.8)}\\
& {\sysname} & 62.6 {\tiny(5.7)} / 63.1 {\tiny(6.7)} & 62.1 {\tiny(4.1)}\\
\midrule

\multirow{2}{*}{$|\mathcal{K}| = 20$}
& BERT-base-Classic &33.1 {\tiny(1.9)} / 33.4 {\tiny(2.0)} & 49.5 {\tiny(5.4)} \\
& RoBERTa-base-Classic  & 36.1 {\tiny(1.4)} / 36.5 {\tiny(1.4)} & 51.9 {\tiny(4.5)}\\
& RoBERTa-large-Classic & 35.8 {\tiny(1.0)} / 36.8 {\tiny(1.5)} & 51.0 {\tiny(4.8)}   \\
& BERT-base-Prompt  & 42.8 {\tiny(2.1)} / 44.5 {\tiny(2.8)} & 50.5 {\tiny(3.1)}
 \\
 & RoBERTa-base-Prompt &51.9 {\tiny(2.9)} / 52.8 {\tiny(3.1)} & 57.5 {\tiny(3.4)}  \\

& RoBERTa-large-Prompt & 60.3 {\tiny(2.0)} / 61.6 {\tiny(2.7)} & 63.0 {\tiny(2.4)}\\
& {\sysname} &70.3 {\tiny(4.0)} / 71.9 {\tiny(4.4)} & 68.2 {\tiny(3.6)}\\
\midrule

\multirow{2}{*}{$|\mathcal{K}| = 30$}
& BERT-base-Classic  &   34.3 {\tiny(2.0)} / 34.5 {\tiny(1.9)} & 51.6 {\tiny(3.8)} \\
& RoBERTa-base-Classic &   38.2 {\tiny(1.9)} / 38.6 {\tiny(2.2)} & 53.1 {\tiny(2.4)} \\
& RoBERTa-large-Classic  &   38.0 {\tiny(1.7)} / 39.0 {\tiny(3.1)} & 51.4 {\tiny(3.7)} \\
& BERT-base-Prompt  &  44.7 {\tiny(2.4)} / 45.7 {\tiny(2.4)} & 52.6 {\tiny(4.0)}
 \\
 & RoBERTa-base-Prompt &  53.6 {\tiny(2.4)} / 55.0 {\tiny(3.0)} & 61.0 {\tiny(4.7)} \\

& RoBERTa-large-Prompt  &   62.8 {\tiny(2.6)} / 64.1 {\tiny(3.3)} & 66.1 {\tiny(2.2)} \\
& {\sysname} &  73.5 {\tiny(2.8)} / {75.0 {\tiny(3.7)}} &71.0 {\tiny(2.4)} \\
\midrule

\multirow{2}{*}{$|\mathcal{K}| = 100$}
& BERT-base-Classic  &  41.6 {\tiny(3.5)} / 42.8 {\tiny(3.3)} & 54.0 {\tiny(3.4)}  \\
& RoBERTa-base-Classic &   45.3 {\tiny(0.9)} / 46.8 {\tiny(0.8)}
 &55.6 {\tiny(5.0)}\\
& RoBERTa-large-Classic  &   49.1 {\tiny(6.6)} / 51.5 {\tiny(6.7)} & 56.8 {\tiny(4.9)}\\
& BERT-base-Prompt   &47.7 {\tiny(1.9)} / 49.8 {\tiny(1.7)} & 52.0 {\tiny(3.3)}
 \\
 & RoBERTa-base-Prompt  & 59.7 {\tiny(1.3)} / 61.3 {\tiny(1.4)} & 64.3 {\tiny(2.2)}\\

& RoBERTa-large-Prompt   & 69.5 {\tiny(1.7)} / 70.9 {\tiny(2.0)} & 72.3 {\tiny(2.9)} \\
& {\sysname} &  78.6 {\tiny(2.4)} / 79.9 {\tiny(1.6)} & 74.3 {\tiny(2.2)}  \\
\midrule

\multirow{2}{*}{$|\mathcal{K}| = 500$}
& BERT-base-Classic    & 52.4 {\tiny(3.7)} / 53.9 {\tiny(3.6)} & 59.2 {\tiny(2.3)}\\
& RoBERTa-base-Classic & 61.3 {\tiny(2.1)} / 63.4 {\tiny(1.8)} & 62.7 {\tiny(7.5)}
 \\
& RoBERTa-large-Classic     & 73.9 {\tiny(1.8)} / 75.6 {\tiny(1.5)} & 66.8 {\tiny(4.9)}\\
& BERT-base-Prompt   & 54.9 {\tiny(0.8)} / 57.6 {\tiny(1.1)} & 57.0 {\tiny(1.6)}\\
& RoBERTa-base-Prompt&69.3 {\tiny(0.6)} / 70.3 {\tiny(0.5)} & 69.5 {\tiny(2.1)}\\
& RoBERTa-large-Prompt      & 78.8 {\tiny(0.8)} / 80.0 {\tiny(0.6)} & 78.2 {\tiny(0.5)}\\
& {\sysname} & 81.9 {\tiny(0.6)} / 82.8 {\tiny(0.6)} & 81.9 {\tiny(1.1)} \\
\midrule

\multirow{2}{*}{$|\mathcal{K}| = 1000$}

& BERT-base-Classic  &   57.4 {\tiny(2.6)} / 59.3 {\tiny(2.2)} & 60.4 {\tiny(3.2)} \\
& RoBERTa-base-Classic &  68.9 {\tiny(0.9)} / 70.2 {\tiny(0.8)} & 66.8 {\tiny(2.9)}\\
& RoBERTa-large-Classic     & 79.0 {\tiny(0.9)} / 80.2 {\tiny(0.8)} & 77.0 {\tiny(1.7)}\\
& BERT-base-Prompt  & 58.9 {\tiny(1.0)} / 61.2 {\tiny(1.0)}&60.5 {\tiny(1.7)} \\
& RoBERTa-base-Prompt &73.5 {\tiny(0.9)} / 74.4 {\tiny(1.1)} & 73.9 {\tiny(1.1)}\\
& RoBERTa-large-Prompt    & 81.6 {\tiny(1.0)} / 82.6 {\tiny(0.5)} & 78.5 {\tiny(1.8)}\\
& {\sysname} & 83.9 {\tiny(0.8)} / 84.6 {\tiny(0.5)} & 82.9 {\tiny(1.5)}\\
\midrule
\end{tabular}
}
\caption{Average performance and standard deviation of different encoders with Classic and Prompt-tuning strategies with various training labels
$|\mathcal{K}|$. }
\label{tab_app:var_shot_model}
\end{table*}

\begin{table*}[h]
\small
\centering
\resizebox{\textwidth}{!}{
\begin{tabular}{clllllllll}
\toprule
\textbf{Labels} &  \textbf{Tuning} & \textbf{\#Params} &  \textbf{Avg} & {\textbf{MNLI (m/mm)}}  & {\textbf{RTE}}   & {\textbf{QQP}} & {\textbf{SST-2}} & {\textbf{Subj}} & {\textbf{MPQA}} \\
&&&&(acc) &(acc) &(acc) & (acc)& (acc)&(acc)\\
\midrule

\multirow{4}{*}{$|\mathcal{K}| = 10$}
& Full  & 355M&69.3&54.8 {\tiny(3.7)} / 55.6 {\tiny(4.6)} & 60.0 {\tiny(4.4)} & 58.7 {\tiny(4.6)} & 89.5 {\tiny(1.7)} & 84.5 {\tiny(8.6)} & 67.8 {\tiny(6.9)}\\\cdashline{2-10}

& Embedding & 53M&62.3&53.3 {\tiny(1.1)} / 53.7 {\tiny(1.2)}& 56.1 {\tiny(3.5)} & 50.9 {\tiny(6.4)}  &84.4 {\tiny(3.6)}&70.3 {\tiny(6.0)}&58.8 {\tiny(7.0)} \\
& Attention & 101M& 68.0 &55.1 {\tiny(3.0)} / 55.8 {\tiny(4.0)} & 57.9 {\tiny(3.9)}& 57.8 {\tiny(7.0)}  & 90.3 {\tiny(1.5)} &82.0 {\tiny(6.6)} &64.3 {\tiny(6.6)}
\\
& FF-output & 102M&69.0&55.7 {\tiny(3.3)} / 56.4 {\tiny(4.0)} &60.4 {\tiny(4.3)} & 59.1 {\tiny(5.7)} & 90.2 {\tiny(1.5)} & 82.2 {\tiny(7.1)} & 66.2 {\tiny(8.1)} \\
& FF-intermediate &102M &67.1&55.0 {\tiny(2.8)} / 55.7 {\tiny(3.7)} & 57.7 {\tiny(3.5)} & 57.0 {\tiny(7.2)} 
& 89.3 {\tiny(2.1)} & 80.7 {\tiny(6.1)} & 62.7 {\tiny(6.9)} \\

\midrule

\multirow{4}{*}{$|\mathcal{K}| = 20$}

& Full &355M &75.4&60.3 {\tiny(2.0)} / 61.6 {\tiny(2.7)} & 64.3 {\tiny(2.4)} & 67.8 {\tiny(4.2)}  & 90.6 {\tiny(1.8)} &  88.3 {\tiny(2.2)} & 80.6 {\tiny(7.5)} \\\cdashline{2-10}
& Embedding & 53M&65.6&53.2 {\tiny(1.3)} / 53.1 {\tiny(1.5)} & 58.1 {\tiny(0.9)} & 55.7 {\tiny(5.2)} & 86.0 {\tiny(1.7)} & 78.0 {\tiny(2.0)} & 62.7 {\tiny(3.2)} \\
& Attention &101M & 74.6&59.2 {\tiny(1.7)} / 60.2 {\tiny(2.4)} &  61.4 {\tiny(2.2)} & 66.8 {\tiny(2.6)}  & 91.7 {\tiny(1.1)} &  88.6 {\tiny(1.5)} & 79.3 {\tiny(5.5)}
\\

& FF-output& 102M&75.7&60.2 {\tiny(1.8)} / 61.4 {\tiny(2.6)} & 65.2 {\tiny(2.5)} & 67.7 {\tiny(3.4)} & 91.4 {\tiny(1.4)} &  88.5 {\tiny(1.3)} & 80.3 {\tiny(5.2)}

 \\
& FF-intermediate &102M &73.5&58.3 {\tiny(1.6)} / 59.3 {\tiny(2.0)} & 60.8 {\tiny(2.3)} & 66.2 {\tiny(3.2)}  & 90.5 {\tiny(1.3)} & 87.4 {\tiny(2.3)} & 77.4 {\tiny(5.8)} \\

\midrule

\multirow{4}{*}{$|\mathcal{K}| = 30$}

& Full & 355M &77.6&62.8 {\tiny(2.6)} / 64.1 {\tiny(3.3)} & 66.1 {\tiny(2.2)} &  71.1 {\tiny(1.5)}  & 91.5 {\tiny(1.0)}& 91.0 {\tiny(0.5)} &82.7 {\tiny(3.8)}\\\cdashline{2-10}
& Embedding &53M &67.0&54.1 {\tiny(1.1)} / 54.0 {\tiny(1.2)} & 59.0 {\tiny(2.7)}  & 56.7 {\tiny(4.5)}  &  85.8 {\tiny(0.9)} & 82.2 {\tiny(2.6)} & 64.2 {\tiny(2.1)} \\
& Attention &101M & 77.0 &61.6 {\tiny(2.2)} / 62.7 {\tiny(2.9)} & 65.8 {\tiny(3.2)} & 70.1 {\tiny(2.2)}  & 91.7 {\tiny(0.9)} & 90.4 {\tiny(0.7)} & 82.1 {\tiny(2.5)}\\
& FF-output & 102M&77.6&62.3 {\tiny(2.1)} / 63.5 {\tiny(3.0)} & 67.3 {\tiny(2.6)} &  70.8 {\tiny(1.7)}  & 91.8 {\tiny(0.8)} & 90.2 {\tiny(1.3)} & 82.5 {\tiny(3.4)}

\\
& FF-intermediate  &102M & 75.9& 60.4 {\tiny(1.9)} / 61.4 {\tiny(2.5)} & 64.0 {\tiny(3.9)} &  69.0 {\tiny(2.7)}  & 91.0 {\tiny(1.2)} & 90.0 {\tiny(1.3)} &80.7 {\tiny(2.7)}

\\

\midrule

\end{tabular}
}
\caption{Average performance and standard deviation on tuning different modules of RoBERTa-large with varying amount of training labels $|\mathcal{K}|$.}
\label{tab_app:adapter_pilot}
\end{table*}

\begin{table*}[h]
\small
\centering
\resizebox{\textwidth}{!}{
\begin{tabular}{llclllllll}
\toprule
 \textbf{Tuning} & \textbf{\#Params} & \multicolumn{1}{c}{\textbf{MNLI (m/mm)}}  & \multicolumn{1}{c}{\textbf{RTE}}   & \multicolumn{1}{c}{\textbf{QQP}} & \multicolumn{1}{c}{\textbf{SST-2}} & \multicolumn{1}{c}{\textbf{Subj}} & \multicolumn{1}{c}{\textbf{MPQA}} \\
\midrule

 Head-only &1M&54.1 {\tiny(1.1)} / 54.1 {\tiny(1.3)} & 58.8 {\tiny(2.6)} & 56.7 {\tiny(4.5)} & 85.6 {\tiny(1.0)}& 82.1 {\tiny(2.5)} & 64.1 {\tiny(2.1)} \\
 Bias-only& 1M & 54.4 {\tiny(1.3)} / 54.4 {\tiny(1.5)}& 59.8 \tiny{(3.5)} & 58.6 {\tiny(4.4)}  & 87.3 {\tiny(1.1)} & 83.9 {\tiny(2.3)}& 65.8 {\tiny(1.8)}  \\
 
 Prompt-only &1M & 47.3 {\tiny(0.2)} / 47.7 {\tiny(0.1)} & 53.0 {\tiny(0.6)} &39.9 {\tiny(0.7)}  & 75.7 {\tiny(1.7)} & 51.5 {\tiny(1.4)} & 70.9 {\tiny(2.4)}
 
  \\
  {\sysname} Adapter (2)& 1M & 56.3 {\tiny(3.8)} / 57.1 {\tiny(4.7)} & 63.7 {\tiny(4.9)} & 68.2 {\tiny(2.4)} & 89.2 {\tiny(0.9)} & 90.2 {\tiny(0.8)} & 68.4 {\tiny(3.0)}\\
 
  Houlsby Adapter &14M & 35.7 {\tiny(1.1)} / 36.2 {\tiny(2.0)} & 51.0 {\tiny(3.0)} & 62.8 {\tiny(3.0)}  & 57.0 {\tiny(6.2)} & 83.2 {\tiny(5.4)} & 57.2 {\tiny(3.5)}

\\
 {\sysname} Adapter (128)  & 14M & \textbf{62.4 {\tiny(1.7)} / 63.7 {\tiny(2.5)}} &\textbf{66.6 {\tiny(3.9)}}&\textbf{71.2 {\tiny(2.6)} }&\textbf{91.7 {\tiny(1.0)}} & \textbf{90.9 {\tiny(1.3)}} & \textbf{82.6 {\tiny(2.0)}}
 \\ 
 \midrule
  Full tuning & 355M  &62.8 {\tiny(2.6)} / 64.1 {\tiny(3.3)} & 66.1 {\tiny(2.2)} &  71.1 {\tiny(1.5)}  & 91.5 {\tiny(1.0)}& 91.0 {\tiny(0.5)} &82.7 {\tiny(3.8)}\\

\midrule
 
\end{tabular}
}
\caption{Average performance and standard deviation of several lightweight parameter-efficient prompt-tuning strategies with $|\mathcal{K}| = 30$ training labels. The best performance is shown in \textbf{bold} along with the number ($\#$) of adapter parameters of total encoder parameters. }

\label{tab_app:adapter}
\end{table*}

{\textbf{Comparisons over different PLMs.} Table~\ref{tab_app:var_encoder_30}, \ref{tab_app:var_encoder_20} and \ref{tab_app:var_encoder_10} show the performance comparison of two representative PLMs with different parameters using prompt-based FN on 10, 20 and 30 labeled samples.  We observe that average performance increases with increase in model size within each model family. Overall, we observe RoBERTa models to perform much better than BERT. This observation is consistent with the observation in Table~\ref{tab:var_model}.

}
\begin{table*}[h]
\small
\centering
\resizebox{0.5\textwidth}{!}{
\begin{tabular}{llc}
\toprule
\textbf{Backbone}      & \textbf{Approach}     & \textbf{Average Acc}\\
\midrule
BERT-base    & Prompt FN & 66.0\\
BERT-base   & MetaST & 60.2\\
BERT-base  & PromptST & 66.1\\
BERT-base  & LiST &68.6\\
\midrule
BERT-large  & Prompt FN & 67.0\\
BERT-large & MetaST & 60.1\\
BERT-large& PromptST&67.6\\
BERT-large & LiST & 70.6\\
\midrule
RoBERTa-base    & Prompt FN & 73.0\\
RoBERTa-base & MetaST & 62.9\\
RoBERTa-base& PromptST& 73.1\\
RoBERTa-base& LiST & 76.4\\
\midrule
RoBERTa-large    & Prompt FN & 77.6\\
RoBERTa-large  & MetaST& 62.6\\
RoBERTa-large & PromptST&77.2\\
RoBERTa-large & LiST & 82.0\\
\midrule
\end{tabular}
}
\caption{{Average performance over various backbones  with  with training labels $|K|=30$ (with unlabeled data). MetaST, PromptST and LiST are semi-supervised approaches. }}

\label{tab_app:var_encoder_30}
\end{table*}

\begin{table*}[h]
\small
\centering
\resizebox{0.5\textwidth}{!}{
\begin{tabular}{llc}
\toprule
\textbf{Backbone}      & \textbf{Approach}     & \textbf{Average Acc}\\

\midrule
BERT-base    & Prompt FN &64.4\\
BERT-base   & MetaST & 57.7\\
BERT-base  & PromptST& 64.9\\
BERT-base  & LiST & 66.5\\
\midrule
BERT-large  & Prompt FN & 64.8\\
BERT-large & MetaST & 57.7\\
BERT-large& PromptST&65.6\\
BERT-large & LiST & 68.5\\
\midrule
RoBERTa-base    & Prompt FN & 71.2\\
RoBERTa-base & MetaST &59.8\\
RoBERTa-base& PromptST&71.5\\
RoBERTa-base& LiST & 75.1\\
\midrule
RoBERTa-large    & Prompt FN &75.4\\
RoBERTa-large  & MetaST & 58.9\\
RoBERTa-large & PromptST& 74.8\\
 RoBERTa-large & LiST & 79.5\\
 \midrule
\end{tabular}
}
\caption{{Average performance over various backbones  with  with training labels $|K|=20$ (with unlabeled data). MetaST, PromptST and LiST are semi-supervised approaches.}}

\label{tab_app:var_encoder_20}
\end{table*}

\begin{table*}[h]
\small
\centering
\resizebox{0.5\textwidth}{!}{
\begin{tabular}{llc}
\toprule
\textbf{Backbone}      & \textbf{Approach}     & \textbf{Average Acc}\\

\midrule
BERT-base    & Prompt FN & 58.2\\
BERT-base   & MetaST & 52.4\\
BERT-base  &  PromptST & 59.6\\
BERT-base  & LiST & 60.9\\
\midrule
BERT-large  & Prompt FN & 59.4\\
BERT-large & MetaST & 53.8\\
BERT-large& PromptST& 59.6 \\
BERT-large & LiST &  62.1\\
\midrule
RoBERTa-base    & Prompt FN & 66.8\\
RoBERTa-base & MetaST & 54.1\\
RoBERTa-base& PromptST& 66.5\\
RoBERTa-base& LiST&  69.4\\
\midrule
RoBERTa-large    & Prompt FN & 69.3\\
RoBERTa-large  & MetaST& 53.8\\
RoBERTa-large & PromptST& 68.2\\
RoBERTa-large & LiST& 72.8\\
\midrule
\end{tabular}
}
\caption{{ Average performance over various backbones  with  with training labels $|K|=10$ (with unlabeled data). MetaST, PromptST and LiST are semi-supervised approaches.}}

\label{tab_app:var_encoder_10}
\end{table*}

{\textbf{More ablation Analysis.}  Tables~\ref{tab_app:ablation_30}, ~\ref{tab_app:ablation_20} and ~\ref{tab_app:ablation_10} show the performance of LiST (14 MM parameters) by removing different components as well as LiST without (w/o) adapter (355 MM parameters). It can be observed  that the trend is consistent over different shots.  ``w/o re-init``  leads to performance drop consistently in various shots and different data sets.  Adapter with 4\% tunable parameters obtains similar performance to full model tuning for shots of 10, 20 and 30 as shown in Table 8. 
}

\begin{table*}[h]
\small
\centering
\resizebox{0.6\textwidth}{!}{
\begin{tabular}{lllc}
\toprule
&MNLI & RTE\\
\midrule
LIST         &   73.5(2.8) / 75.0(3.7) &71.0(2.4)\\
w/o re-init  &  66.7(2.8) / 68.3(4.3) & 69.0(4.9)\\
w/o re-weighting &72.9(3.4) / 74.2(4.5) & 69.7(4.1)\\
w/o warmup & 67.9(12.9) / 69.0(13.1) & 69.2(4.5)\\
w/ hard pseudo-labels & 71.7(3.8) / 73.0(5.4) & 69.5(4.2) \\
\midrule
w/o Adapter (Full Model) &73.6(2.7) / 74.8(2.7) & 71.2(2.3) \\
\midrule
\end{tabular}
}
\caption{{ Ablation analysis of LiST with \# of training data = 30.}}

\label{tab_app:ablation_30}
\end{table*}

\begin{table*}[h]
\small
\centering
\resizebox{0.6\textwidth}{!}{
\begin{tabular}{lllc}
\toprule
&MNLI & RTE\\
\midrule
LiST &        71.8(2.3) / 73.0(3.1)     &  69.0(3.5)\\
w/o re-init  &  65.6(2.6) / 66.9(3.4) & 66.5(3.7)\\
w/o re-weighitng&70.7(4.1) / 71.8(4.6) & 67.1 (5.6)\\
w/o warmup& 66.9(5.4) / 68.3(5.7) & 67.4(5.1)\\
w/ hard pseudo labels & 69.9(3.6) / 71.4(3.7) & 67.7(3.5)\\
\midrule
w/o Adapter (Full Model) & 66.6 (3.2) / 68.1 (3.6) & 69.69 (5.29)\\
\midrule
\end{tabular}
}
\caption{ {Ablation analysis of LiST with \# of training data = 20.} }

\label{tab_app:ablation_20}
\end{table*}

\begin{table*}[h]
\small
\centering
\resizebox{0.6\textwidth}{!}{
\begin{tabular}{lllc}
\toprule
&MNLI & RTE\\
\midrule
LiST & 65.0(4.5) / 66.3(4.9) & 64.2(2.8)\\
w/o re-init  &  58.7(4.4) / 59.4(5.5)   & 58.8(4.0)\\
w/o re-weighting & 63.8(5.8) / 64.5(6.6) & 61.7(2.6)\\
w/o warmup & 62.7(5.2) / 63.3(6.2) & 61.7(4.8)\\
w/ hard pseudo labels & 60.8(6.6) / 61.8 (6.8) & 60.8(3.1)\\
\midrule
w/o Adapter (Full model) & 60.0 (3.7) / 61.1 (4.8)& 62.4 (6.79)\\
\midrule
\end{tabular}
}
\caption{{ Ablation analysis of LiST with \# of training data = 10.}}

\label{tab_app:ablation_10}
\end{table*}

\begin{table*}[h]
\small
\centering
\resizebox{\textwidth}{!}{
\begin{tabular}{clllllllll}
\toprule
\textbf{Labels} & \textbf{Models} & {\textbf{Avg}} & \textbf{\#Tunable} &\multicolumn{1}{l}{\textbf{MNLI (m/mm)}}  & \multicolumn{1}{l}{\textbf{RTE}}   & \multicolumn{1}{l}{\textbf{QQP}} & \multicolumn{1}{l}{\textbf{SST-2}} & \multicolumn{1}{l}{\textbf{Subj}} & \multicolumn{1}{l}{\textbf{MPQA}} \\
&&&{\bf Params}&(acc) &(acc) &(acc) & (acc)& (acc)&(acc)\\
\midrule

\multirow{1}{*}{\thead{$|\mathcal{K}| = 30$ }}

& Classic FN &60.9 &355M &38.0 {\tiny(1.7)} / 39.0 \tiny{(3.1)} & 51.4 {\tiny(3.7)} &64.3 {\tiny(8.1)}  & 65.0 {\tiny(11.5)} & 90.2 {\tiny(2.2)} &56.1 {\tiny(5.3)}\\

\midrule
\multirow{1}{*}{\thead{$|\mathcal{K}| = 30$   +Unlabeled Data}}
& LIST w/ Classic FN & 66.7   &14M    & \textbf{39.9  {\tiny(5.6)} / 41.7  {\tiny(7.6)}}  &  \textbf{54.9  {\tiny(1.4)}} & \textbf{67.4  {\tiny(7.0)}} & \textbf{73.6  {\tiny(9.9)}} &  \textbf{92.3  {\tiny(1.1)}} & \textbf{71.4  {\tiny(4.7)}} \\
 \midrule
 
\end{tabular}
}\vspace{-0.1in}
\caption{{Performance comparison of classic FN with RoBERTa-large as the encoder with standard deviation in parantheses.  The best performance is shown in \textbf{bold}.}}
\label{tab:different_fn}
\vspace{-0.18in}
\end{table*}

{\textbf{Adapters w/ different number of training labels.} We compare the performance of LiST Adapter (14 MM parameters) against full model tuning (355 MM parameters) where we obtain 96\% tunable parameter reduction with almost matching performance across 10, 20 and 30 shots in Table~\ref{tab_app:adapter_var_shots}.
}

\begin{table*}[h]
\small
\centering
\resizebox{0.5\textwidth}{!}{
\begin{tabular}{llc}
\toprule
 \# of Training data & Approach     & Average Acc (Six Tasks)\\
\midrule 
 30& Full tuning & 77.6\\
30 & LiST Adapter & 77.7\\
\midrule 
 20& Full tuning & 75.4\\
20 & LiST Adapter & 75.2\\
\midrule 
10& Full tuning & 69.3\\
10 & LiST Adapter & 68.9\\
\midrule

\end{tabular}
}
\caption{{Average Accuracy of Adapter w/ various number of training labels (No Semi-supervised Setting).}}
\label{tab_app:adapter_var_shots}

\end{table*}

\end{document}